\documentclass{article}

\usepackage{microtype}
\usepackage{graphicx}
\usepackage{booktabs} % for professional tables

\usepackage{hyperref}

\usepackage[accepted]{icml2020}

\usepackage{xr-hyper}
\usepackage{amsmath}
\usepackage{amsfonts}
\usepackage{color}
\usepackage{dsfont,graphicx}
\usepackage{subfig}
\usepackage{booktabs}
\usepackage{bm}
\usepackage{paralist}
\usepackage{mathtools}
\usepackage{amsthm}
\theoremstyle{plain} % default

\newtheorem{theorem}{Theorem}

\newtheorem{definition}{Definition}

\newcommand{\fracpartial}[2]{\frac{\partial #1}{\partial  #2}}
\icmltitlerunning{The Tree Ensemble Layer: Differentiability meets Conditional Computation}

\begin{document}

\twocolumn[
\icmltitle{The Tree Ensemble Layer: Differentiability meets Conditional Computation}

% It is OKAY to include author information, even for blind
% submissions: the style file will automatically remove it for you
% unless you've provided the [accepted] option to the icml2020
% package.

% List of affiliations: The first argument should be a (short)
% identifier you will use later to specify author affiliations
% Academic affiliations should list Department, University, City, Region, Country
% Industry affiliations should list Company, City, Region, Country

% You can specify symbols, otherwise they are numbered in order.
% Ideally, you should not use this facility. Affiliations will be numbered
% in order of appearance and this is the preferred way.
\icmlsetsymbol{equal}{*}

\begin{icmlauthorlist}
\icmlauthor{Hussein Hazimeh}{mit}
\icmlauthor{Natalia Ponomareva}{google}
\icmlauthor{Petros Mol}{google}
\icmlauthor{Zhenyu Tan}{googlebr}
\icmlauthor{Rahul Mazumder}{mit}
\end{icmlauthorlist}

\icmlaffiliation{mit}{Massachusetts Institute of Technology}
\icmlaffiliation{google}{Google Research}
\icmlaffiliation{googlebr}{Google Brain}

% \icmlaffiliation{goo}{Googol ShallowMind, New London, Michigan, USA}
% \icmlaffiliation{ed}{School of Computation, University of Edenborrow, Edenborrow, United Kingdom}

\icmlcorrespondingauthor{Hussein Hazimeh}{hazimeh@mit.edu}

% You may provide any keywords that you
% find helpful for describing your paper; these are used to populate
% the "keywords" metadata in the PDF but will not be shown in the document
\icmlkeywords{Decision Trees, Tree Ensembles, Neural Networks, Neural Trees, Gradient Boosted Decision Trees, Conditional Computation, Soft Trees, Differentiable Trees}

\vskip 0.3in
]

% this must go after the closing bracket ] following \twocolumn[ ...

% This command actually creates the footnote in the first column
% listing the affiliations and the copyright notice.
% The command takes one argument, which is text to display at the start of the footnote.
% The \icmlEqualContribution command is standard text for equal contribution.
% Remove it (just {}) if you do not need this facility.

\printAffiliationsAndNotice{}  % leave blank if no need to mention equal contribution
%\printAffiliationsAndNotice{\icmlEqualContribution} % otherwise use the standard text.

\begin{abstract}
Neural networks and tree ensembles are state-of-the-art learners, each with its unique statistical and computational advantages. We aim to combine these advantages by introducing a new layer for neural networks, composed of an ensemble of differentiable decision trees (a.k.a. soft trees). While differentiable trees demonstrate promising results in the literature, they are typically slow in training and inference as they do not support conditional computation. We mitigate this issue by introducing a new sparse activation function for sample routing, and implement true conditional computation by developing specialized forward and backward propagation algorithms that exploit sparsity. Our efficient algorithms pave the way for jointly training over deep and wide tree ensembles using first-order methods (e.g., SGD). Experiments on 23 classification datasets indicate over $10$x speed-ups compared to the differentiable trees used in the literature and over $20$x reduction in the number of parameters compared to gradient boosted trees, while maintaining competitive performance.
%improvements over gradient boosted trees in terms of predictive performance and compactness. 
Moreover, experiments on CIFAR, MNIST, and Fashion MNIST indicate that replacing dense layers in CNNs with our tree layer reduces the  test loss by $7$-$53\%$ and the number of parameters by $8$x. We provide an open-source TensorFlow implementation with a Keras API.
\end{abstract}

\section{Introduction}
Decision tree ensembles have proven very successful in various machine learning applications. Indeed, they are often referred to as the best ``off-the-shelf'' learners \cite{friedman2001elements}, as they exhibit several appealing properties such as ease of tuning, robustness to outliers, and interpretability \cite{friedman2001elements, xgboost}. Another natural property in trees is \textsl{conditional computation}, which refers to their ability to route each sample through a small  number of nodes (specifically, a single root-to-leaf path). Conditional computation can be broadly defined as the ability of a model to activate only a small part of its architecture in an input-dependent fashion \cite{DBLP:journals/corr/BengioBPP15}. This can lead to both computational benefits and enhanced statistical properties. On the computation front, routing samples through a small part of the tree leads to substantial training and inference speed-ups compared to methods that do not route samples. Statistically, conditional computation offers the flexibility to reduce the number of parameters used by each sample, which can act as a regularizer \cite{Breiman1983ClassificationAR, friedman2001elements, DBLP:journals/corr/BengioBPP15}. 

However, the performance of trees relies on feature engineering, since they lack a good mechanism for representation learning \cite{bengio2013representation}. This is an area in which neural networks (NNs) excel, especially in speech and image recognition applications \cite{bengio2013representation, he2015delving, yu2016automatic}. However, NNs do not naturally support conditional computation and are harder to tune.

In this work, we combine the advantages of neural networks and tree ensembles by designing a hybrid model.
Specifically, we propose the \textsl{Tree Ensemble Layer (TEL)} for neural networks. This layer is an additive model of differentiable decision trees, can be inserted anywhere in a neural network, and is trained along with the rest of the network using gradient-based optimization methods (e.g., SGD). While differentiable trees in the literature show promising results, especially in the context of neural networks, e.g., \citet{criminisi2016deep, FrosstH17}, they do not offer true conditional computation. We equip TEL with a novel mechanism to perform conditional computation, during both training and inference. We make this possible by introducing a new sparse activation function for sample routing, along with specialized forward and backward propagation algorithms that exploit sparsity. Experiments on 23 real datasets indicate that TEL achieves over $10$x speed-ups compared to the current differentiable trees, without sacrificing predictive performance. %In particular, we introduce a new differentiable activation function which allows for routing samples in exactly one direction (left or right). Moreover, we design specialized forward and backward propagation algorithms that exploit sparsity to enable conditional computation.

Our algorithms pave the way for jointly optimizing over both wide and deep tree ensembles. Here joint optimization refers to updating all the trees simultaneously (e.g., using first-order methods like SGD). This has been a major computational challenge prior to our work. For example, jointly optimizing over classical (non-differentiable) decision trees is a hard combinatorial problem \cite{friedman2001elements}. Even with differentiable trees, the training complexity grows exponentially with the tree depth, making joint optimization difficult \cite{criminisi2016deep}. A common approach is to train tree ensembles using greedy ``stage-wise'' procedures, where only one tree is updated at a time and never updated again---this is a main principle in gradient boosted decision trees (GBDT) \cite{friedman2001greedy}\footnote{\textcolor{black}{There are follow-up works on GBDT which update the leaves of all trees simultaneously, e.g., see \citet{johnson2013learning}. However, our approach allows for updating both the internal node and leaf weights simultaneously.}}. We hypothesize that joint optimization yields more compact and expressive ensembles than GBDT. Our experiments confirm this, indicating that TEL can achieve over $20$x reduction in model size. This can have important implications for interpretability, latency and storage requirements during inference. 

\textbf{Contributions: } Our contributions can be summarized as follows: \textbf{(i)} We design a new differentiable activation function for trees which allows for routing samples through small parts of the tree (similar to classical trees).
\textbf{(ii)} We realize conditional computation by developing specialized forward and backward propagation algorithms that exploit sparsity to achieve an optimal time complexity. Notably, the complexity of our backward pass can be independent of the tree depth and is generally better than that of the forward pass---this is not possible in backpropagation for neural networks.  \textbf{(iii)} We perform experiments on a collection of 26 real datasets, which confirm TEL as a competitive alternative to current differentiable trees, GBDT, and dense layers in CNNs.  \textbf{(iv)} We provide an open-source TensorFlow implementation of TEL along with a Keras interface\footnote{\url{https://github.com/google-research/google-research/tree/master/tf_trees} \label{footnote:url}}.
\paragraph{Related Work: }

Table \ref{table:related} summarizes the most relevant related work.
\begin{table}[htbp]
\centering
\caption{Related work on conditional computation}
\label{table:related}
\footnotesize
\begin{tabular}{@{}lllll@{}}
\toprule
Paper   & CT  & CI & DO &  Model/Optim          \\ \midrule
\citet{criminisi2016deep} & N & N & Y & Soft tree/Alter \\
\citet{ioannou2016decision} & N & H & Y & Tree-NN/SGD \\
\citet{FrosstH17} & N & H & Y & Soft tree/SGD \\
\citet{DBLP:journals/corr/ZoranLB17} & N & H & N & Soft tree/Alter \\
%\cite{pmlr-v70-jernite17a} & N & N & N & Soft tree/SGD \\
\citet{DBLP:journals/corr/ShazeerMMDLHD17} & H & Y &  N & Tree-NN/SGD \\
\citet{tanno2018adaptive} & N & H & Y & Soft tree/SGD \\
\citet{Biau2019} & H & N & Y & Tree-NN/SGD \\
\citet{hehn2019end} & N & H & Y & Soft tree/SGD \\
Our method & Y & Y & Y & Soft tree/SGD \\
\bottomrule
\multicolumn{5}{l}{
\footnotesize 
\parbox[t]{.94\linewidth}{
\textit{H} is heuristic (e.g., training model is different from  inference), %\textit{CT} is conditional (as opposed to dense) training that explores only a sparse subset of model weights. 
\textit{CT} is conditional training. \textit{CI} is conditional inference. \textit{DO} indicates whether the objective function is differentiable. \textit{Soft tree} refers to a differentiable tree, whereas \textit{Tree-NN} refers to NNs with a tree-like structure. %\textit{Optim} describes whether the learning is done end-to-end (via SGD for example) or in alternating fashion (switching between learning trees and feature representation).
\textit{Optim} stands for optimization (SGD or alternating minimization).
}}
\end{tabular}
% \raggedright{\textit{H} is heuristic, \textit{CT} is conditional (as opposed to dense) training that explores only a sparse subset of model weights.  \textit{CI} is conditional inference. \textit{DL} indicates whether the loss being optimized is truly differentiable. \textit{Soft tree} refers to a differentiable tree, whereas \textit{Tree-NN} refers to NNs stacked in a tree-like structure. \textit{Optim} describes whether the learning is done end-to-end (via SGD for example) or in alternating fashion (switching between learning trees and feature representation).}
\end{table}
\raggedbottom
\textcolor{black}{Differentiable decision trees (a.k.a. soft trees) are an instance of the Hierarchical Mixture of Experts introduced by \citet{jordan1994hierarchical}}. The internal nodes of these trees act as routers, sending samples to the left and right with different proportions. This framework does not support conditional computation as each sample is processed in all the tree nodes. Our work avoids this issue by allowing each sample to be routed through small parts of the tree, without losing differentiability. A number of recent works have used soft trees in the context of deep learning. For example, \citet{criminisi2016deep} equipped soft trees with neural representations and used alternating minimization to learn the feature representations and the leaf outputs. \citet{hehn2019end} extended \citet{criminisi2016deep}'s approach to allow for conditional inference and growing trees level-by-level. \citet{FrosstH17} trained a (single) soft tree using SGD and leveraged a deep neural network to expand the dataset used in training the tree.  %whereas \cite{criminisi2016deep}, \cite{DBLP:journals/corr/ZoranLB17} used alternating minimization to learn the feature representations and the leaf outputs. 
\citet{DBLP:journals/corr/ZoranLB17} also leveraged a tree structure with a routing mechanism similar to soft trees, in order to equip the k-nearest neighbors algorithm with neural representations. All of these works have observed that computation in a soft tree can be expensive. Thus, in practice, heuristics are used to speed up inference, e.g., \citet{FrosstH17} uses the root-to-leaf path with the highest probability during inference, leading to discrepancy between the models used in training and inference. 
%However, this leads to a discrepancy between the models used in training and inference. 
%To avoid dense computations during training, a number of papers employed various heuristics. \cite{FrosstH17}  introduced a temperature by which the output of the split node is scaled before passing through the sigmoid. Such soft trees tended to get stuck in poor solutions areas when internal nodes were hard routing the examples, and thus becoming impossible to update due to the gradients becoming zero. They mitigated it by introducing a regularizer that encouraged 50-50 splitting for internal nodes. The strength of this regularizer had to be depth dependent, because encouraging deeper nodes to do equal split was hurting the performance.
%During inference, a number of works use only the path with the highest probability or a heuristic chosen (\cite{FrosstH17} \cite{DBLP:journals/corr/ZoranLB17}), which makes the inference procedure different from training, thus potentially hurting the predictive performance.
Instead of making a tree differentiable, \citet{pmlr-v70-jernite17a} hypothesized about properties the best tree should have, and introduced a pseudo-objective that encourages balanced and pure splits. %They optimized this objective with SGD along with some intermediate steps to choose the labels assignments in the leaves. 
They optimized using SGD along with intermediate processing steps. %to choose the leaf outputs. 

Another line of work introduces tree-like structure to NNs via some routing mechanism. For example, \citet{ioannou2016decision} employed tree-shaped CNNs with branches as weight matrices with sparse block diagonal structure. \citet{DBLP:journals/corr/ShazeerMMDLHD17} created the Sparsely-Gated Mixture-of-Experts layer where samples are routed to subnetworks selected by a trainable gating network. %\cite{Biau2019} used the fact that any decision tree can be represented as 3 layer neural network and proposed to learn the structure of the tree using CART, then reformulate it as a neural net and fine tune the coefficients using SGD.
\citet{Biau2019} represented a decision tree using a 3-layer neural network and combined CART and SGD for training.
\citet{tanno2018adaptive}
looked into adaptively growing an NN with routing nodes for performing tree-like conditional computations. %Adaptive growth happened by choosing, using the validation set, between the actions of adding a router, a non-linear transformation or moving onto the next node. 
%For inference, \cite{ioannou2016decision, DBLP:journals/corr/ShazeerMMDLHD17, tanno2018adaptive} routed samples to the top \textit{k} branches only, where \textit{k} was either a hyperparameter or 1. Such inference is different from the training inference and the objective is not truly differentiable.
However, in these works, the inference model is either different from training or the router is not differentiable (but still trained using SGD)---see Table \ref{table:related} for details.

% \section{Contributions}
% Our contributions are as follows:

\section{The Tree Ensemble Layer} \label{section:tree}
TEL is an additive model of differentiable decision trees. In this section, we introduce TEL formally and then discuss the routing mechanism used in our trees.
For simplicity, we assume that TEL is used as a standalone layer. %(i.e., without any other layers). 
Training trees with other layers will be discussed in Section \ref{sec:conditional}. 

We assume a supervised learning setting, with input space $\mathcal{X} \subseteq \mathbb{R}^p$ and output space $\mathcal{Y} \subseteq \mathbb{R}^{k}$. For example, in the case of regression (with a single output) $k=1$, while in classification $k$ depends on the number of classes.  Let $m$ be the number of trees in the ensemble, and let $T^{(j)}: \mathcal{X} \to \mathbb{R}^{k}$ be the $j$th tree in the ensemble. For an input sample $x \in \mathbb{R}^{p}$, the output of the layer is a sum over all the tree outputs:
\begin{align} \label{eq:additivetrees}
\mathcal{T}(x) = T^{(1)} (x) + T^{(2)} (x) + \dots + T^{(m)} (x).
\end{align}
\textcolor{black}{The output of the layer, $\mathcal{T}(x)$, is a vector in $\mathbb{R}^{k}$ containing raw predictions}. In the case of classification, mapping from raw predictions to $\mathcal{Y}$ can be done by applying a softmax and returning the class with the highest probability. Next, we introduce the key building block of the approach: the differentiable decision tree.

\textbf{The Differentiable Decision Tree}:
Classical decision trees perform \textsl{hard routing}, i.e., a sample is routed to exactly one direction at every internal node. Hard routing introduces discontinuities in the loss function, making trees unamenable to continuous optimization. Therefore, trees are usually built in a greedy fashion. 
%In this section, we introduce a differentiable variant of decision trees known as soft trees: these trees perform \textsl{soft routing}, where every internal node can route the sample to the left and right simultaneously with different proportions. This routing mechanism makes soft trees differentiable, so learning a tree can be done using gradient-based methods. Soft trees used in the literature cannot route a sample only to the left or to the right, making conditional computation impossible. Subsequently, we introduce soft trees along with a new activation function, which allows conditional computation while preserving differentiability. 
In this section, we present an enhancement of the soft trees proposed by \citet{jordan1994hierarchical} and utilized in \citet{criminisi2016deep, FrosstH17, hehn2019end}. Soft trees are a variant of decision trees that perform \textsl{soft routing}, where every internal node can route the sample to the left and right simultaneously, with different proportions. This routing mechanism makes soft trees differentiable, so learning can be done using gradient-based methods. Soft trees cannot route a sample exclusively to the left or to the right, making conditional computation impossible. Subsequently, we introduce a new activation function for soft trees, which allows conditional computation while preserving differentiability.

We consider a single tree in the additive model \eqref{eq:additivetrees}, and denote the tree by $T$ (we drop the superscript to simplify the notation). Recall that $T$ takes an input sample and returns an output vector (logit), i.e., $T: \mathcal{X} \subseteq \mathbb{R}^{p} \to \mathbb{R}^{k}$. Moreover, we assume that $T$ is a perfect binary tree with depth $d$. We use the sets $\mathcal{I}$ and $\mathcal{L}$ to denote the internal (split) nodes and the leaves of the tree, respectively. %(i.e., the set of all nodes in the path starting from the root and ending at the parent of node $i$). 
For any node $i \in \mathcal{I} \cup \mathcal{L}$, we define $\mathcal{A}(i)$ as its set of ancestors and use the notation $\{ x \to i \}$ for the event that a sample $x \in \mathbb{R}^{p}$ reaches $i$. \textcolor{black}{A summary of the notation used in this paper can be found in Table A.1 in the appendix. }

\textbf{Soft Routing: } Internal tree nodes perform soft routing, where a sample is routed left and right with different proportions. We will introduce soft routing using a probabilistic model. While we use probability to model the routing process, we will see that the final prediction of the tree is an expectation over the leaves, making $T$ a deterministic function. \textcolor{black}{Unlike classical decision trees which use axis-aligned splits, soft trees are based on hyperplane (a.k.a. oblique) splits \citep{10.5555/1622826.1622827}, where a linear combination of the features is used in making routing decisions. Particularly, each internal node $i \in \mathcal{I}$ is associated with a trainable weight vector $w_i \in \mathbb{R}^{p}$ that defines the node's hyperplane split}. Let $\mathcal{S}: \mathbb{R} \to [0,1]$ be an activation function. Given a sample $x \in \mathbb{R}^{p}$, the probability that internal node $i$ routes $x$ to the left is defined by $\mathcal{S}(\langle w_i, x \rangle)$.

Now we discuss how to model the probability that $x$ reaches a certain leaf $l$. %Let $\{ x \to l \}$ denote  the event that sample $x$ reaches leaf $l \in \mathcal{L}$. 
Let $[l \swarrow i]$ (resp. $[i \searrow l]$) denote the event that leaf $l$ belongs to the left (resp. right) subtree of node $i \in \mathcal{I}$. Assuming that the routing decision made at each internal node in the tree is independent of the other nodes, the probability that $x$ reaches $l$ is given by:
\begin{align} \label{eq:probxtol}
P(\{ x \to l \}) = \prod\nolimits_{i \in \mathcal{A}(l)} r_{i,l}(x),
\end{align}
where $r_{i,l}(x)$ is the probability of node $i$ routing $x$ towards the subtree containing leaf $l$, i.e.,
% \begin{align} \label{eq:r}
%  r_{i,l}(x) := \mathcal{S}(\langle x, w_i \rangle ) ^{\mathds{1}[l \swarrow i]} (1 - \mathcal{S}(\langle x, w_i \rangle))^{\mathds{1}[i \searrow l]}
% \end{align}
$
 {r_{i,l}(x) := \mathcal{S}(\langle x, w_i \rangle ) ^{\mathds{1}[l \swarrow i]} (1 - \mathcal{S}(\langle x, w_i \rangle))^{\mathds{1}[i \searrow l]}}.
$
%where we recall that $\mathcal{A}(l)$ is the set of ancestors of leaf $l$. 
Next, we define how the root-to-leaf probabilities in \eqref{eq:probxtol} can be used to make the final prediction of the tree. 

\textbf{Prediction: }
As with classical decision trees, we assume that each leaf stores a weight vector $o_{l} \in \mathbb{R}^{k}$ (learned during training). Note that, during a forward pass, $o_l$ is a constant vector, meaning that it is not a function of the input sample(s). For a sample $x \in \mathbb{R}^p$, we define the prediction of the tree as the expected value of the leaf outputs, i.e., 
\begin{align} \label{eq:T}
 T(x) = \sum\nolimits_{l \in \mathcal{L}} P(\{ x \to l \}) o_{l}.
\end{align}
%where $P(\{ x \to l \})$ is defined in \eqref{eq:probxtol}.
\textbf{Activation Functions: }
In soft routing, the internal nodes use an activation function  $\mathcal{S}$ in order to compute the routing probabilities. The logistic (a.k.a. sigmoid) function is the common choice for $\mathcal{S}$ in the literature on soft trees (see \citet{jordan1994hierarchical, criminisi2016deep, FrosstH17, tanno2018adaptive, hehn2019end}). While the logistic function can output arbitrarily small values, it cannot output an exact zero. This implies that any sample $x$ will reach every node in the tree with a positive probability (as evident from \eqref{eq:probxtol}). Thus, computing the output of the tree in \eqref{eq:T} will require computation over every node in the tree, an operation which is exponential in tree depth. 

\begin{figure}[htbp]
\centering
\captionsetup[subfloat]{farskip=0pt} % subfloat adds extra space for subcaptions, but we're not using subcaptions
\includegraphics[trim={1cm 0.5cm 0.5cm 1cm},clip, width=6.2cm]{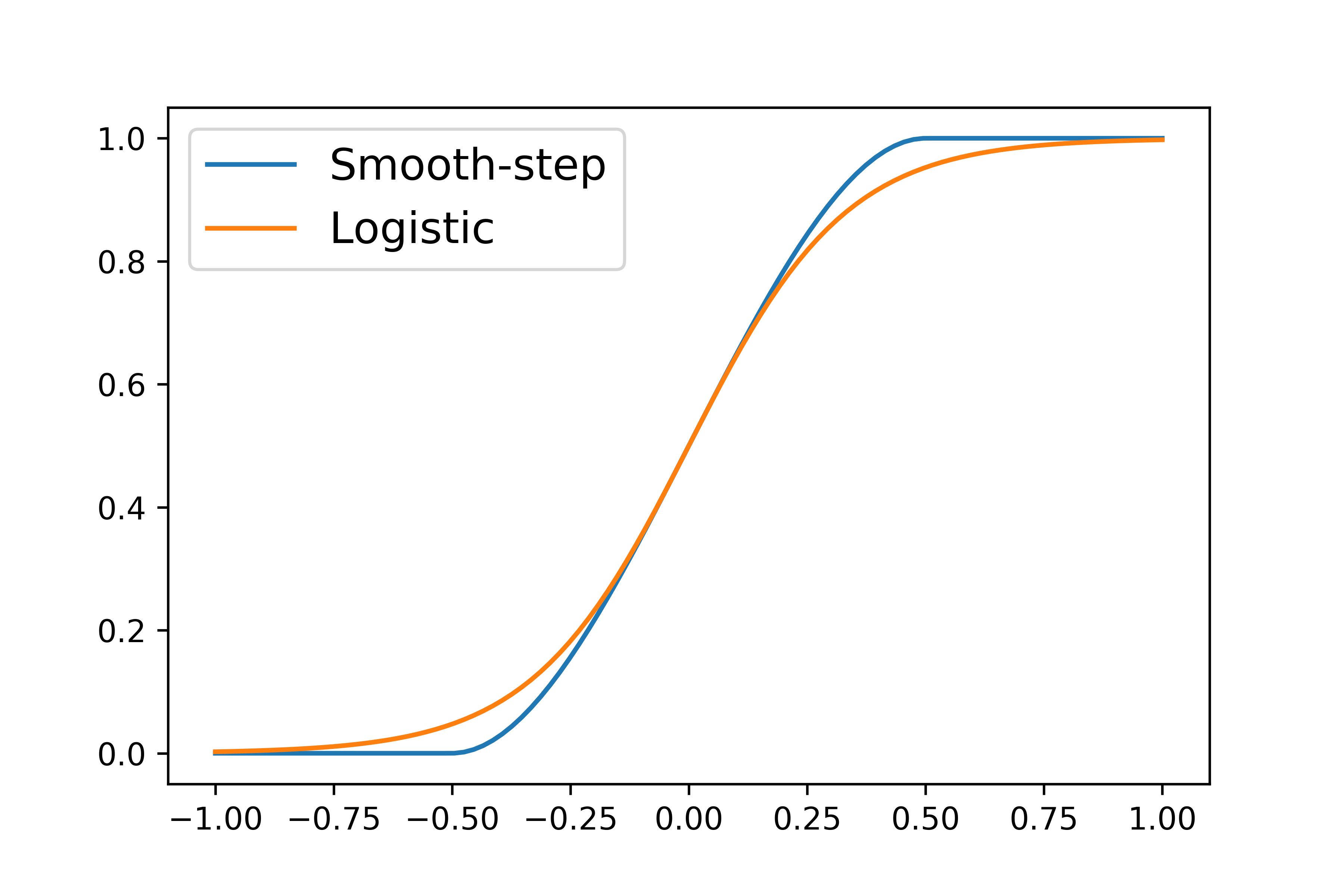}
\caption{%A plot of the smooth-step function in \eqref{eq:smoothstep} with $\gamma=1$ and the logistic function defined by $(1 + e^{-6t})^{-1}$.
Smooth-step vs. Logistic $(1 + e^{-6t})^{-1}$.}
\label{fig:smoothstep}
\end{figure}

We propose a novel \textit{smooth-step activation function}, which can output exact zeros and ones, thus allowing for true conditional computation. Our smooth-step function is S-shaped and continuously differentiable, similar to the logistic function. Let $\gamma$ be a non-negative scalar parameter. The smooth-step function is a cubic polynomial in the interval $[-\gamma/2,\gamma/2]$, $0$ to the left of the interval, and $1$ to the right. More formally, we assume that the function takes the parametric form $\mathcal{S}(t) = a t^3 + b t^2 + ct + d$ for $t \in [-\gamma/2,\gamma/2]$, where $a,b,c,d$ are scalar parameters. We then solve for the parameters under the following continuity and differentiability constraints: (i) $\mathcal{S}(-\gamma/2) = 0$, (ii) $\mathcal{S}(\gamma/2) = 1$, (iii) $\mathcal{S}'(t)|_{t=-\gamma/2} =\mathcal{S}'(t)|_{t=\gamma/2} = 0$. This leads to:
\begin{align} \label{eq:smoothstep}
    \mathcal{S}(t) = 
        \begin{cases}
            0 & \text{ if } t \leq -\gamma/2 \\
            -\frac{2}{\gamma^{3}}t^3 + \frac{3}{2\gamma}t + \frac{1}{2} & \text{ if } -\gamma/2 \leq t \leq \gamma/2 \\
            1 & \text{ if } t \geq \gamma/2
        \end{cases}
\end{align}
By construction, the smooth-step function in \eqref{eq:smoothstep} is continuously differentiable for any $t \in \mathbb{R}$ (including $-\gamma/2$ and $\gamma/2$). In Figure \ref{fig:smoothstep}, we plot the smooth-step (with $\gamma=1$) and logistic activation functions; the logistic function here takes the form $(1 + e^{-6t})^{-1}$, i.e., it is a rescaled variant of the standard logistic function, so that the two functions are on similar scales. The two functions can be very close in the middle of the fractional region. The main difference is that the smooth-step function outputs exact zero and one, whereas the logistic function converges to these asymptotically. 
% \begin{figure}[htbp]
% \centering
% \includegraphics[scale=0.35]{Figures/actplot.png}
% \caption{%A plot of the smooth-step function in \eqref{eq:smoothstep} with $\gamma=1$ and the logistic function defined by $(1 + e^{-6t})^{-1}$.
% Smooth-step vs. the logistic function $(1 + e^{-6t})^{-1}$.}
% \label{fig:smoothstep}
% \end{figure}

% \begin{figure}[htbp]
% \centering
% \captionsetup[subfloat]{farskip=0pt} % subfloat adds extra space for subcaptions, but we're not using subcaptions
% \subfloat{\raisebox{1ex}{\includegraphics[trim={1cm 0.5cm 0.5cm 1cm},clip, width=4.25cm]{Figures/actplot.png}}} 
% \hspace{-0.4cm} 
% \subfloat{\includegraphics[trim={0cm 0.2cm 0.2cm 0.3cm},clip, width=4.25cm]{Figures/reachable.png}}
% \caption{%A plot of the smooth-step function in \eqref{eq:smoothstep} with $\gamma=1$ and the logistic function defined by $(1 + e^{-6t})^{-1}$.
% \textbf{Left}: Smooth-step vs. Logistic $(1 + e^{-6t})^{-1}$. \textbf{Right}: Number of reachable leaves (per sample) during training on a tree of depth $10$.}
% \label{fig:smoothstep}
% \end{figure}
Outside $[-\gamma/2, \gamma/2]$, the smooth-step function performs hard routing, similar to classical decision trees. The choice of $\gamma$ controls the fraction of samples that are hard routed. A very small $\gamma$ can lead to many zero gradients in the internal nodes, whereas a very large $\gamma$ might limit the extent of conditional computation. In our experiments, we use batch normalization \cite{ioffe2015} before the tree layer so that the inputs to the smooth-step function remain centered and bounded. This turns out to be very effective in preventing the internal nodes from having zero gradients, at least in the first few training epochs. Moreover, we view $\gamma$ as a hyperparameter, which we tune over the range $[10^{-4},1]$. This range works well for balancing the training performance and conditional computation across the 26 datasets we used (see Section \ref{label-experiments}). 

\begin{figure}[htbp]
\centering
\captionsetup[subfloat]{farskip=0pt} % subfloat adds extra space for subcaptions, but we're not using subcaptions
\includegraphics[trim={0cm 0.2cm 0.2cm 0.3cm},clip, width=6.2cm]{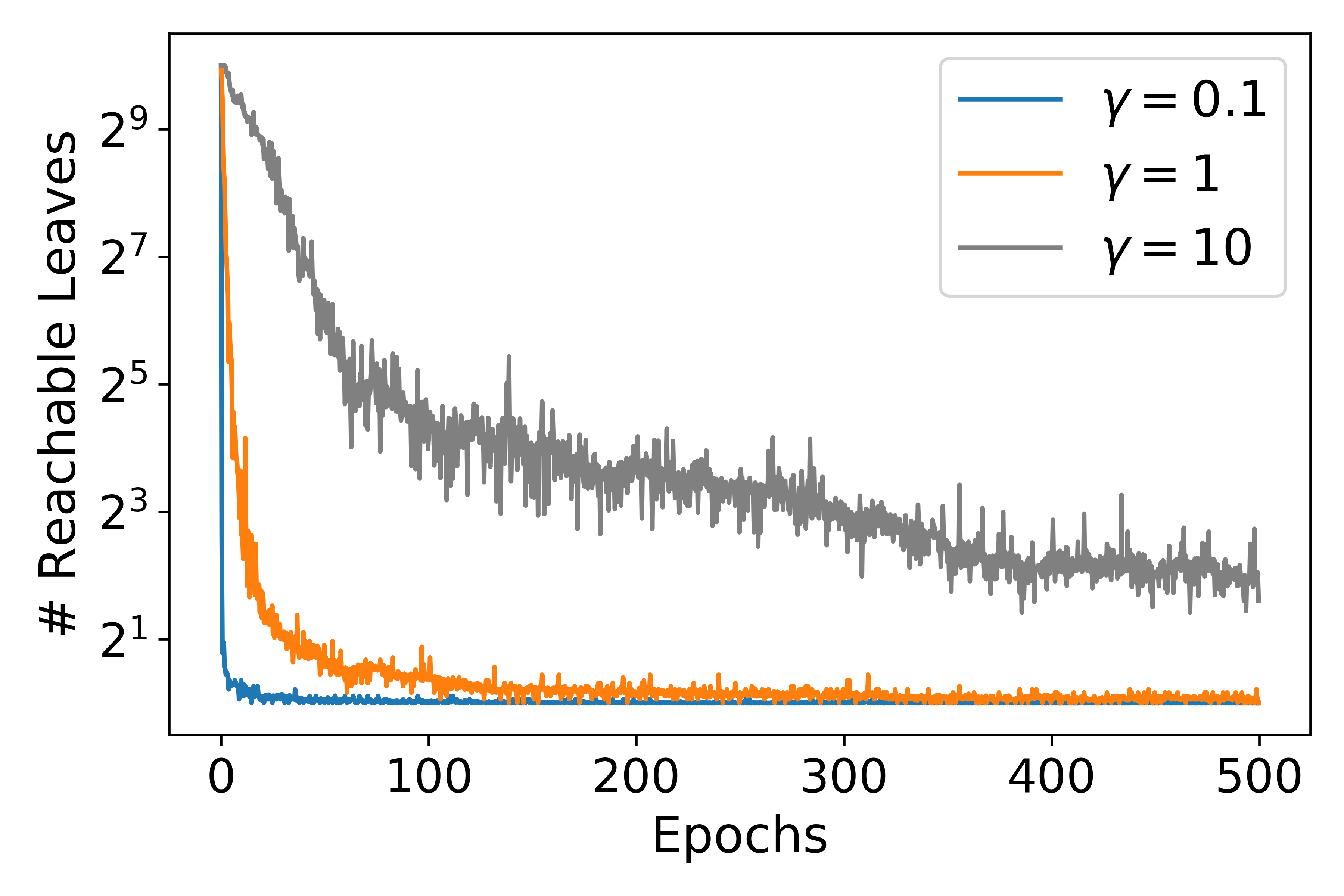}
\caption{Number of reachable leaves (per sample) during training a tree of depth $10$.}
\label{fig:reachable_leaves}
\end{figure}

For a given sample $x$, we say that a node $i$ is reachable if $P(x \to i ) > 0$. The number of reachable leaves directly controls the extent of conditional computation. In Figure \ref{fig:reachable_leaves}, we plot the average number of reachable leaves (per sample) as a function of the training epochs, for a single tree of depth $10$ (i.e., with $1024$ leaves) and different $\gamma$'s. This is for the diabetes dataset \cite{Olson2017PMLB}, using Adam \cite{kingma2014adam} for optimization (see the appendix for details). The figure shows that for small enough $\gamma$ (e.g., $\gamma \le 1$), the number of reachable leaves rapidly converges to $1$ during training (note that the y-axis is on a log scale). We observed this behavior on all the  datasets in our experiments. 

\textcolor{black}{We note that variants of the smooth-step function are popular in computer graphics \citep{ebert2003texturing,rost2009opengl}. However, to our knowledge, the smooth-step function has not been used in soft trees or neural networks. It is also worth mentioning that the cubic polynomial used for interpolation in \eqref{eq:smoothstep} can be substituted with higher-order polynomials (e.g, polynomial of degree 5, where the first and second derivatives vanish at $\pm \gamma/2$). The algorithms we propose in Section \ref{sec:conditional} directly apply to the case of higher-order polynomials.}

In the next section, we show how the sparsity in the smooth-step function and in its gradient can be exploited to develop efficient forward and backward propagation algorithms.

\section{Conditional Computation} \label{sec:conditional}
We propose using first-order optimization methods (e.g., SGD and its variants) to optimize TEL. A main computational bottleneck in this case is the gradient computation, whose time and memory complexities can grow exponentially in the tree depth. This has hindered training large tree ensembles in the literature. In this section, we develop efficient forward and backward propagation algorithms for TEL by exploiting the sparsity in both the smooth-step function and its gradient. We show that our algorithms have optimal time complexity and discuss cases where they run significantly faster than standard backpropagation.

\textbf{Setup: } We assume a general setting where TEL is a hidden layer. %, i.e., it is preceded and succeeded by other layers. %, all of which we assume to be  sub-differentiable. 
Without loss of generality, we consider only one sample  and one tree. Let $x \in \mathbb{R}^{p}$ be the input to TEL and denote the tree output by $T(x) \in \mathbb{R}^{k}$, where $T(x)$ is defined in \eqref{eq:T}.  We use the same notation as in Section \ref{section:tree}, and we collect the leaf vectors $o_l$, $l\in\mathcal{L}$ into the matrix $O \in \mathbb{R}^{|\mathcal{L}| \times k}$ and the internal node weights $w_i$, $i \in \mathcal{I}$ into the matrix $W \in \mathbb{R}^{|\mathcal{I}| \times p}$. Moreover, for a differentiable function $h(z)$ which maps $\mathbb{R}^{s} \to \mathbb{R}^{u}$, we denote its Jacobian by $\fracpartial{h}{z} \in \mathbb{R}^{u \times s}$. Let $L$ be the loss function to be optimized (e.g., cross-entropy). Our goal is to efficiently compute the following three gradients: $\fracpartial{L}{O}$, $\fracpartial{L}{W}$, and $\fracpartial{L}{x}$. The first two gradients are needed by the optimizer to update $O$ and $W$. The third gradient is used to continue the backpropagation in the layers preceding TEL. We assume that a backpropagation algorithm has already computed the gradients associated with the layers after TEL and has computed $\fracpartial{L}{T}$.

\textbf{Number of Reachable Nodes: } To exploit conditional computation effectively, each sample should reach a relatively small number of leaves. This can be enforced by choosing the parameter $\gamma$ of the smooth-step function to be sufficiently small. When analyzing the complexity of the forward and backward passes below, we will assume that the sample $x$ reaches $U$ leaves and $N$ internal nodes.

% \textbf{Comparison with standard trees: } Note that our soft trees differ from standard decision trees in several ways. Firstly, each split is a hyperplane split, taking into account all the features, as opposed to one feature or a group of features split as in standard decision trees. Thus, the trees are essentially oblique \cite{10.5555/1622826.1622827}. Additionally, constants on leaves are also optimized during the learning via SGD, as opposed to being a function over the instances that end up in this leaf, as in standard Decision Trees.

\subsection{Conditional Forward Pass} Prior to computing the gradients, a forward pass over the tree is required. This entails computing expression \eqref{eq:T}, which is a sum of probabilities over all the root-to-leaf paths in $T$. Our algorithm exploits the following observation: if a certain edge on the path to leaf $l$ has a zero probability, then $P(x \to l )=0$ so there is no need to continue evaluation along that path. Thus, we traverse the tree starting from the root, and every time a node outputs a $0$ probability on one side, we ignore all of its descendants lying on that side. The summation in \eqref{eq:T} is then performed only over the leaves reached by the traversal. We present the conditional forward pass in Algorithm 1, where for any internal node $i$, we denote the left and right children by $left(i)$ and $right(i)$. 
\begin{algorithm}[htbp]
%   \footnotesize
  \caption{Conditional Forward Pass}
  \label{alg:forward}
\begin{algorithmic}[1]
  \STATE {\bfseries Input:} Sample $x \in \mathbb{R}^p$ and tree parameters $W$ and $O$.
  \STATE {\bfseries Output:} $T(x)$
  \STATE \COMMENT{For any node $i$, $i.prob$ denotes $P(x \to i)$.}
  \STATE \COMMENT{$to\_traverse$ is a stack for traversing nodes.}
  \STATE $output \gets 0$, $to\_traverse \gets \{ root \}$, $root.prob \gets 1$
  \WHILE{$to\_traverse$ is not empty }
  \STATE Remove a node $i$ from $to\_traverse$
%  \STATE Compute and store $P(\{ x \to i \})$ 
  \IF{$i$ is an internal node}
  \STATE $left(i).prob = i.prob * \mathcal{S}(\langle w_i, x \rangle)$ 
  \STATE $right(i).prob = i.prob * (1-\mathcal{S}(\langle w_i, x \rangle))$ 
  %\STATE Compute and store $\langle w_i, x \rangle$ and $\mathcal{S}_{\gamma}(\langle w_i, x \rangle)$
  %\STATE Add to $to\_traverse$ $left(i)$ if $\mathcal{S}(\langle w_i, x \rangle) = 1$, $right(i)$ if $\mathcal{S}(\langle w_i, x \rangle) = 0$, and $left(i)$ and $right(i)$ o.w.
  \STATE if $\mathcal{S}(\langle w_i, x \rangle) > 0$, add $left(i)$ to $to\_traverse$
  \STATE if $\mathcal{S}(\langle w_i, x \rangle) < 1$, add $right(i)$ to $to\_traverse$
    %   \IF{$\mathcal{S}(\langle w_i, x \rangle) = 0$}
    %   \STATE $right(i).prob = i.prob * (1-\mathcal{S}(\langle w_i, x \rangle)) $
    %   \STATE Add $right(i)$ to $to\_traverse$
    %   \ELSIF{$\mathcal{S}(\langle w_i, x \rangle) = 1$}
    %   \STATE $left(i).prob = i.prob * \mathcal{S}(\langle w_i, x \rangle)$
    %   \STATE Add $left(i)$ to $to\_traverse$
    %   \ELSE
    %   \STATE $left(i).prob = i.prob * \mathcal{S}(\langle w_i, x \rangle)$
    %   \STATE $right(i).prob = i.prob * (1-\mathcal{S}(\langle w_i, x \rangle)) $
    %   \STATE Add $left(i)$ and $right(i)$ to $to\_traverse$
    %   \ENDIF
  \ELSE
  \STATE $output \gets output + i.prob * o_i$
  \ENDIF
  \ENDWHILE
\end{algorithmic}
\end{algorithm}
\raggedbottom
\\
\textbf{Time Complexity: } The algorithm visits each reachable node in the tree once. Every reachable internal node requires $\mathcal{O}(p)$ operations to compute $\mathcal{S}(\langle w_i, x \rangle)$, whereas each reachable leaf requires $\mathcal{O}(k)$ operations to update the output variable. Thus, the overall complexity is $\mathcal{O}(N p + U k)$ (recall that $N$ and $U$ are the number of reachable internal nodes and leaves, respectively). This is in contrast to a dense forward pass\footnote{By dense forward pass, we mean evaluating the tree without conditional computation (as in a standard forward pass).}, whose complexity is $\mathcal{O}(2^{d} p + 2^{d}k)$ (recall that $d$ is the depth). As long as $\gamma$ is chosen so that $U$ is sub-exponential\footnote{A function $f(t)$ is sub-exp. in $t$ if $\lim_{t \to \infty} {\log(f(t))}/{t} = 0$.} in $d$, the conditional forward pass has a better complexity than the dense pass (this holds since $N = \mathcal{O}(U d)$, implying that $N$ is also sub-exponential in $d$).
% \paragraph{Memory Complexity: } The memory requirements depend on whether the forward pass is being used for inference or training. For inference, a node can be discarded as soon as it is traversed. Since we traverse the tree in a depth-first manner, the worst-case memory complexity is $\mathcal{O}(d)$. When used in the context of training, additional quantities need to be stored in order to perform the backward pass efficiently. In particular, we need to store $l.prob$ for every reachable leaf $l$, and $\langle w_i, x \rangle$ for every ancestor $i$ of the reachable leaves whose  $\mathcal{S}(\langle w_i, x \rangle)$ is fractional---this will be discussed in more detail when we present the backward pass. Note that the number of  these ancestors is equal to $U - 1$ (see the discussion after Definition \ref{def:fractional} for details). Thus, the worst-case memory complexity when used in the context of training is $\mathcal{O}(d + U)$. This is contrast to a standard forward pass, whose complexity is $\mathcal{O}(2^\text{d})$.

\textbf{Memory Complexity: } The memory complexity for inference and training is $\mathcal{O}(d)$ and $\mathcal{O}(d + U)$, respectively. See the appendix for a detailed analysis. This is in contrast to a dense forward pass, whose complexity in training is $\mathcal{O}(2^\text{d})$.
\subsection{Conditional Backward Pass} \label{sec:backward}
Here we develop a backward pass algorithm to efficiently compute the three gradients: 
 $\fracpartial{L}{O}$, $\fracpartial{L}{W}$, and $\fracpartial{L}{x}$, assuming that $\fracpartial{L}{T}$ is available from a backpropagation algorithm. In what follows, we will see that as long as $U$ is sufficiently small, the gradients $\fracpartial{L}{O}$ and $\fracpartial{L}{W}$ will be sparse, and $\fracpartial{L}{x}$ can be computed by considering only a small number of nodes in the tree. 
Let $\mathcal{R}$ be the set of leaves reached by Algorithm 1. The following set turns out to be critical in understanding the sparsity structure in the problem:
${\mathcal{F} := \{i \in \mathcal{I} \ | \  i \in \mathcal{A}(l), \  l \in \mathcal{R}, \  0 < \mathcal{S}(\langle x, w_i \rangle) < 1  \}}$. In words, $\mathcal{F}$ is the set of ancestors of the reachable leaves, whose activation is fractional.

In Theorem 1, we show how the three gradients can be computed by only considering the internal nodes in  $\mathcal{F}$ and leaves in  $\mathcal{R}$. Moreover, the theorem presents sufficient conditions for which the gradients are zero; in particular, $\fracpartial{L}{w_i} = 0$ for every internal node $i \in \mathcal{F}^c$ and  $\fracpartial{L}{o_l} = 0$ for every leaf $l \in \mathcal{R}^{c}$ (where $A^c$ is the complement of a set $A$).
\begin{theorem}
% Define $\mu_1(x,i) = {\fracpartial{\mathcal{S}(t)}{t}|_{t = \langle x, w_i \rangle}}({\mathcal{S}(\langle x, w_i \rangle)})^{-1}$, $\mu_2(x,i) = {\fracpartial{\mathcal{S}(t)}{t}|_{t = \langle x, w_i \rangle}}({(1 - \mathcal{S}(\langle x, w_i \rangle))})^{-1}$,
Define $\mu_1(x,i) = {\fracpartial{\mathcal{S}(\langle x, w_i \rangle)}{\langle x, w_i \rangle}}/{\mathcal{S}(\langle x, w_i \rangle)}$, $\mu_2(x,i) = {\fracpartial{\mathcal{S}(\langle x, w_i \rangle)}{\langle x, w_i \rangle}}/{(1 - \mathcal{S}(\langle x, w_i \rangle))}$,
% Define $\mu_1(x,i) = {\mathcal{S}'(\langle x, w_i \rangle)}/{\mathcal{S}(\langle x, w_i \rangle)}$, $\mu_2(x,i) = {\mathcal{S}'(\langle x, w_i \rangle)}/{(1 - \mathcal{S}(\langle x, w_i \rangle))}$,
and $g(l) =  P(\{ x \to l \}) \langle \fracpartial{L}{T}, o_{l} \rangle$. The gradients needed for backpropagation can be expressed as follows:
 \begin{align*}
% \fracpartial{L}{x} & = \sum_{i \in \mathcal{F}}  \mu_1(x,i) w_i^T \sum_{l \in \mathcal{R} | [l \swarrow i]} g(l) \\
% & - \sum_{i \in \mathcal{F}} \mu_2(x,i)  w_i^T \sum_{l \in \mathcal{R} | [i \searrow l]} g(l)
 & \fracpartial{L}{x} = \sum_{i \in \mathcal{F}}  w_i^T  \Big[ \mu_1(x,i) \sum_{\mathclap{l \in \mathcal{R} | [l \swarrow i]}} g(l) - \mu_2(x,i) \sum_{\mathclap{l \in \mathcal{R} | [i \searrow l]}} g(l) \Big]
\\ 
&  \fracpartial{L}{w_i} = \begin{dcases}
0 & \hspace{-0.3cm} i \in \mathcal{F}^c \\
x^T  \Big[ \mu_1(x,i) \sum_{\mathclap{l \in \mathcal{R} | [l \swarrow i]}} g(l) - \mu_2(x,i) \sum_{\mathclap{l \in \mathcal{R} | [i \searrow l]}} g(l) \Big]& \hspace{-0.2cm} \text{o.w.}
\end{dcases} \\
& \fracpartial{L}{o_l} = \fracpartial{L}{T} P(\{ x \to l \}), \ \forall \ l \in \mathcal{L}
\end{align*}
\end{theorem}
%Theorem 1 is critical for developing can efficient backward pass. In particular, we will see that the theorem allows for computing the gradients with a lower time complexity compared to performing backpropagation on the computational graph corresponding to the reachable tree. %Thus, Theorem 1 can be thought of as a simplification of the computational graph corresponding to the reachable tree.
In Theorem 1, the quantities $\mu_1(x,i)$ and $\mu_2(x,i)$ can be obtained in $\mathcal{O}(1)$ since in Algorithm 1 we store $\langle x, w_i \rangle$ for every $i \in \mathcal{F}$. Moreover, $P(\{ x \to l \})$ is stored in Algorithm 1 for every reachable leaf. However, a direct evaluation of these gradients leads to a  suboptimal time complexity because the terms $\sum_{{l \in \mathcal{R} | [l \swarrow i]}} g(l)$ and $ \sum_{{l \in \mathcal{R} | [i \searrow l]}} g(l)$ will be computed from scratch for every node $i \in \mathcal{F}$. Our conditional backward pass traverses a \textsl{fractional tree}, composed of only the nodes in $\mathcal{F}$ and $\mathcal{R}$, while deploying smart bookkeeping to compute these sums during the traversal and avoid recomputation. We define the {fractional tree} below.
\begin{definition} \label{def:fractional}
Let~$T_{\text{reachable}}$~be the tree traversed by the conditional forward pass (Algorithm 1). We define the fractional tree  $T_{\text{fractional}}$ as the result of the following two operations: (i) remove every internal node $i \in \mathcal{F}^c$ from $T_{\text{reachable}}$ and (ii) connect every node with no parent to its closest ancestor. 
\end{definition}
\textcolor{black}{In Section C.1 of the appendix, we provide an example of how the fractional tree is constructed.} \textcolor{black}{$T_{\text{fractional}}$ is a binary tree with $U$ leaves and $|\mathcal{F}|$ internal nodes, \textcolor{black}{each with exactly 2 children}. It can be readily seen that $|\mathcal{F}| = U - 1$; this relation is useful for analyzing the complexity of the conditional backward pass.} Note that $T_{\text{fractional}}$ can be constructed on-the-fly while performing the conditional forward pass (without affecting its complexity). In Algorithm 2, we present the conditional backward pass, which traverses the fractional tree once and returns $\fracpartial{L}{x}$ and any (potentially) non-zero entries in $\fracpartial{L}{O}$ and  $\fracpartial{L}{W}$.
\begin{algorithm}[htbp]
%   \footnotesize
  \caption{Conditional Backward Pass}
  \label{alg:backward}
\begin{algorithmic}[1]
\STATE {\bfseries Input:} Sample $x \in \mathbb{R}^p$, tree parameters, and $\fracpartial{L}{T}$.
\STATE {\bfseries Output:} $\fracpartial{L}{x}$ and   (potential) non-zeros in $\fracpartial{L}{W}$ and $\fracpartial{L}{O}$.
\STATE $\fracpartial{L}{x} = 0$
%\STATE \COMMENT{For any node $i$, $i.sum\_g$ denotes $\sum_{l \in \mathcal{R} \cap \mathcal{A}(i)} g(l)$}
\STATE \COMMENT{For any node $i$, $i.sum\_g$ denotes $\sum_{l \in \mathcal{R} | i \in \mathcal{A}(l)} g(l)$}
\STATE Traverse $T_{\text{fractional}}$ in post order:
\begin{ALC@g}
\STATE Denote the current node by $i$
\IF {$i$ is a leaf}
\STATE $\fracpartial{L}{o_i} = \fracpartial{L}{T} P(\{ x \to i \})$
\STATE $i.sum\_g = g(i)$ 
\ELSE
\STATE $a =  \mu_1(x,i)$ $(left(i).sum\_g)$
\STATE $b =  \mu_2(x,i)$ $(right(i).sum\_g)$
\STATE $\fracpartial{L}{x} \mathrel{+}=  w_i^T (a - b)$ 
\STATE $\fracpartial{L}{w_i} = x^T (a - b)$
\STATE $i.sum\_g = left(i).sum\_g + right(i).sum\_g$
\ENDIF
\end{ALC@g}
\end{algorithmic}
\end{algorithm}
% \textbf{Time Complexity: } 
% Lines 8 and 9 perform $\mathcal{O}(k)$ operations so each leaf requires $\mathcal{O}(k)$ operations. Lines 11,  12, and 15 are $\mathcal{O}(1)$, while 13 and 14 are $\mathcal{O}(p)$. The total number of internal nodes traversed is $|\mathcal{F}|$. Moreover, we always have $|\mathcal{F}| = U - 1$ (see the discussion following Definition \ref{def:fractional}). Therefore, the worst-case complexity can be written as $\mathcal{O}(Up + Uk)$. In the worst case, Theorem 1 implies that the number of non-zero entries in the three gradients is $p + p|\mathcal{F}| + Uk = \mathcal{O}(Up + Uk)$. Thus, the complexity of the conditional backward pass is optimal, in the sense that its complexity matches the number of non-zero gradient entries, in the worst case. This complexity is generally lower than that of the conditional forward pass whose complexity is $\mathcal{O}(Np + Uk)$. This is because we always have $U = \mathcal{O}(N)$, and there can be many cases where the number of reachable internal nodes $N$ grows faster than $U$. For example, consider a tree with only two reachable nodes ($U=2$) and where the root is the (only) fractional node, then $N$ grows linearly with the depth $d$. Algorithm 2's complexity can be significantly lower than that of a standard backward pass whose complexity is $\mathcal{O}(2^d p + 2^d k)$.

\textbf{Time Complexity: } 
The worst-case complexity of the algorithm is $\mathcal{O}(Up + Uk)$, whereas the best-case complexity is $\mathcal{O}(k)$ (corresponds to $U=1$), and in the worst case, the number of non-zero entries in the three gradients is $\mathcal{O}(Up + Uk)$---see the appendix for analysis. Thus, the complexity is optimal, in the sense that it matches the number of non-zero gradient entries, in the worst case. The worst-case complexity is generally lower than the $\mathcal{O}(Np + Uk)$ complexity of the conditional forward pass. This is because we always have $U = \mathcal{O}(N)$, and there can be many cases where  $N$ grows faster than $U$. For example, consider a tree with only two reachable leaves ($U=2$) and where the root is the (only) fractional node, then $N$ grows linearly with the depth $d$. As long as $U$ is sub-exponential in $d$, Algorithm 2's complexity can be significantly lower than that of a dense backward pass whose complexity is $\mathcal{O}(2^d p + 2^d k)$.

% The best-case complexity of Algorithm 2 is $\mathcal{O}(k)$---this corresponds to the case where there is only one reachable leaf ($U=1$), so the fractional tree is composed of only one node. In our experiments, we observed that this case becomes common after a number of training epochs (the number depends on the learning rate, batch size, and other parameters, but it is typically in the order of tens). 

\textbf{Memory Complexity: } 
We store one scalar per node in the fractional tree (i.e., $i.sum\_g$ for every node $i$ in the fractional tree). Thus, the memory complexity is $\mathcal{O}(|\mathcal{F}| + U) = \mathcal{O}(U)$. If $\gamma$ is chosen so that $U$ is upper-bounded by a constant, then Algorithm 2 will require constant memory. %A standard backward pass can require an exponential memory in the tree depth (the exact memory requirements depend on the specific implementation). 

\textbf{Connections to Backpropagation: }
An interesting observation in our approach is that the conditional backward pass generally has a better time complexity than the conditional forward pass. This is usually impossible in standard backpropagation for NNs, as the forward and backward passes traverse the same computational graph \cite{Goodfellow-et-al-2016}. The improvement in complexity of the backward pass in our case is due to Algorithm 2 operating on the fractional tree, which can contain a significantly smaller number of nodes than the tree traversed by the forward pass. In the language of backpropagation, our fractional tree can be viewed as a ``simplified'' computational graph, where the simplifications are due to Theorem 1.
%approach can be viewed as using two different computational graphs: one for the forward pass and another for the backward pass, where the latter graph is simpler due to exploiting the sparsity in the gradients.

\section{Experiments}
We study the performance of TEL in terms of prediction, conditional computation, and compactness. We evaluate TEL as a standalone learner and as a layer in a NN, and compare to standard soft trees, GBDT, and dense layers.

\textbf{Model Implementation: } TEL is  implemented in TensorFlow 2.0 using custom C++ kernels for forward and backward propagation, along with a Keras Python-accessible interface. \textcolor{black}{The implementation is open source\textsuperscript{\ref{footnote:url}}}.
%\footnote{\url{https://github.com/google-research/google-research/tree/master/tf_trees}}.}

\textbf{Datasets: } We use a collection of 26 classification datasets (binary and multiclass) from various domains (e.g., healthcare, genetics, and image recognition). 23 of these are from the Penn Machine Learning Benchmarks (PMLB) \cite{Olson2017PMLB}, and the 3 remaining are CIFAR-10 \cite{krizhevsky2009learning}, MNIST \cite{lecun1998gradient}, and Fashion MNIST \cite{xiao2017fashion}. Details are in the appendix.

\textbf{Tuning, Toolkits, and Details: } For all the experiments, we tune the hyperparameters using Hyperopt \cite{Bergstra2013} with the Tree-structured Parzen Estimator (TPE). We optimize for either AUC or accuracy with stratified 5-fold cross-validation. NNs (including TEL) were trained using Keras with the TensorFlow backend,  using Adam  \cite{kingma2014adam} and cross-entropy loss. As discussed in Section \ref{section:tree}, TEL is always preceded by a batch normalization layer. GBDT is from XGBoost \cite{xgboost}, Logistic regression and CART are from Scikit-learn \cite{scikit-learn}. Additional details are in the appendix.

\label{label-experiments}
\subsection{Soft Trees: Smooth-step vs. Logistic Activation} \label{sec:softexperiment}
We compare the run time and performance of the smooth-step and  logistic functions using 23 PMLB datasets. 

\textbf{Predictive Performance: } We fix the TEL architecture to 10 trees of depth 4. We tune the learning rate, batch size, and number of epochs (ranges are in the appendix). We assume the following parametric form for the logistic function $f(t) = (1+e^{-t/\alpha})^{-1}$, where $\alpha$ is a hyperparameter which we tune in the range $[10^{-4}, 10^{4}]$. The smooth-step's parameter $\gamma$ is tuned in the range $[10^{-4}, 1]$. Here we restrict the upper range of $\gamma$ to $1$ to enable conditional computation over the whole tuning range. While $\gamma$'s larger than 1 can lead to slightly better predictive performance in some cases, they can slow down training significantly. For tuning, Hyperopt is run for 50 rounds with AUC as the metric. After tuning, models with the best hyperparameters are retrained. We repeat the training procedure 5 times using random weight initializations. The mean test AUC along with its standard error (SE) are in Table \ref{table:activation}.
\begin{table}[htbp]
\footnotesize
\centering
\caption{Test AUC for the smooth-step and logistic functions (fixed TEL architecture). A $*$ indicates statistical significance based on a paired two-sided t-test at a significance level of $0.05$. Best results are in \textbf{bold}. %Results on car-evaluation, dermatology, hyperthyroid, nursery, optdigits, sleep, spambase, texture, twonorm are the same for both functions.}
AUCs on the 9 remaining datasets match and are hence omitted.}
\label{table:activation}
\begin{tabular}{@{}lll@{}}
\toprule
Dataset            & Smooth-step      & Logistic         \\ \midrule
ann-thyroid & $\bm{0.997} \pm 0.0001$ & $0.996 \pm 0.0006$ \\
breast-cancer-w. & $0.992 \pm 0.0015$ & $\bm{0.994} \pm 0.0002$ \\
%car-evaluation & ${1.0} \pm 0.0002$ & ${1.0} \pm 0.0$ \\ %
churn & $0.897 \pm 0.0014$ & $\bm{0.898} \pm 0.0014$ \\
crx & $0.916 \pm 0.0025$ & $\bm{0.929^{*}} \pm 0.0021$ \\
%dermatology & ${0.999} \pm 0.0$ & $0.999 \pm 0.0001$ \\%
diabetes & $\bm{0.832^{*}} \pm 0.0009$ & $0.816 \pm 0.0021$ \\
dna & $0.993 \pm 0.0004$ & $\bm{0.994^{*}} \pm 0.0$ \\
ecoli & $\bm{0.97^{*}} \pm 0.0004$ & $0.952 \pm 0.0038$ \\
flare & $0.78 \pm 0.0027$ & $\bm{0.784} \pm 0.0018$ \\
heart-c & $\bm{0.936} \pm 0.002$ & $0.927 \pm 0.0036$ \\
%hypothyroid & ${0.979} \pm 0.0012$ & ${0.979} \pm 0.0007$ \\%
%nursery & ${0.999} \pm 0.0001$ & ${0.999} \pm 0.0002$ \\%
%optdigits & ${1.0} \pm 0.0001$ & ${1.0} \pm 0.0$ \\%
pima & $\bm{0.828^{*}} \pm 0.0005$ & $0.82 \pm 0.0003$ \\
satimage & $\bm{0.988^{*}} \pm 0.0002$ & $0.987 \pm 0.0002$ \\
%sleep & ${0.922} \pm 0.0001$ & ${0.922} \pm 0.0002$ \\%
solar-flare\_2 & $0.926 \pm 0.0002$ & $\bm{0.927^{*}} \pm 0.0007$ \\
%spambase & ${0.983} \pm 0.0003$ & ${0.983} \pm 0.0003$ \\%
%texture & ${1.0} \pm 0.0$ & $1.0 \pm 0.0$ \\%
%twonorm & ${0.998} \pm 0.0$ & ${0.998} \pm 0.0$ \\%
vehicle & $0.956 \pm 0.0015$ & $\bm{0.965^{*}} \pm 0.0007$ \\
yeast & $\bm{0.876^{*}} \pm 0.0014$ & $0.86 \pm 0.0026$ \\
\midrule
\textit{\# wins} & \textit{7} & \textit{7} \\
\bottomrule
\end{tabular}
\end{table}
\raggedbottom
The smooth-step outperforms the logistic function on 7 datasets (5 are statistically significant). The logistic function also wins on 7 datasets (4 are statistically significant). The two functions match on the rest of the datasets. The differences on the majority of the datasets are small (even when statistically significant), suggesting that using the smooth-step function does not hurt the predictive performance. However, as we will see next, the smooth-step has a significant edge in terms of computation time.

\begin{figure*}[htbp]
    \centering
    \subfloat{{\includegraphics[width=5.8cm]{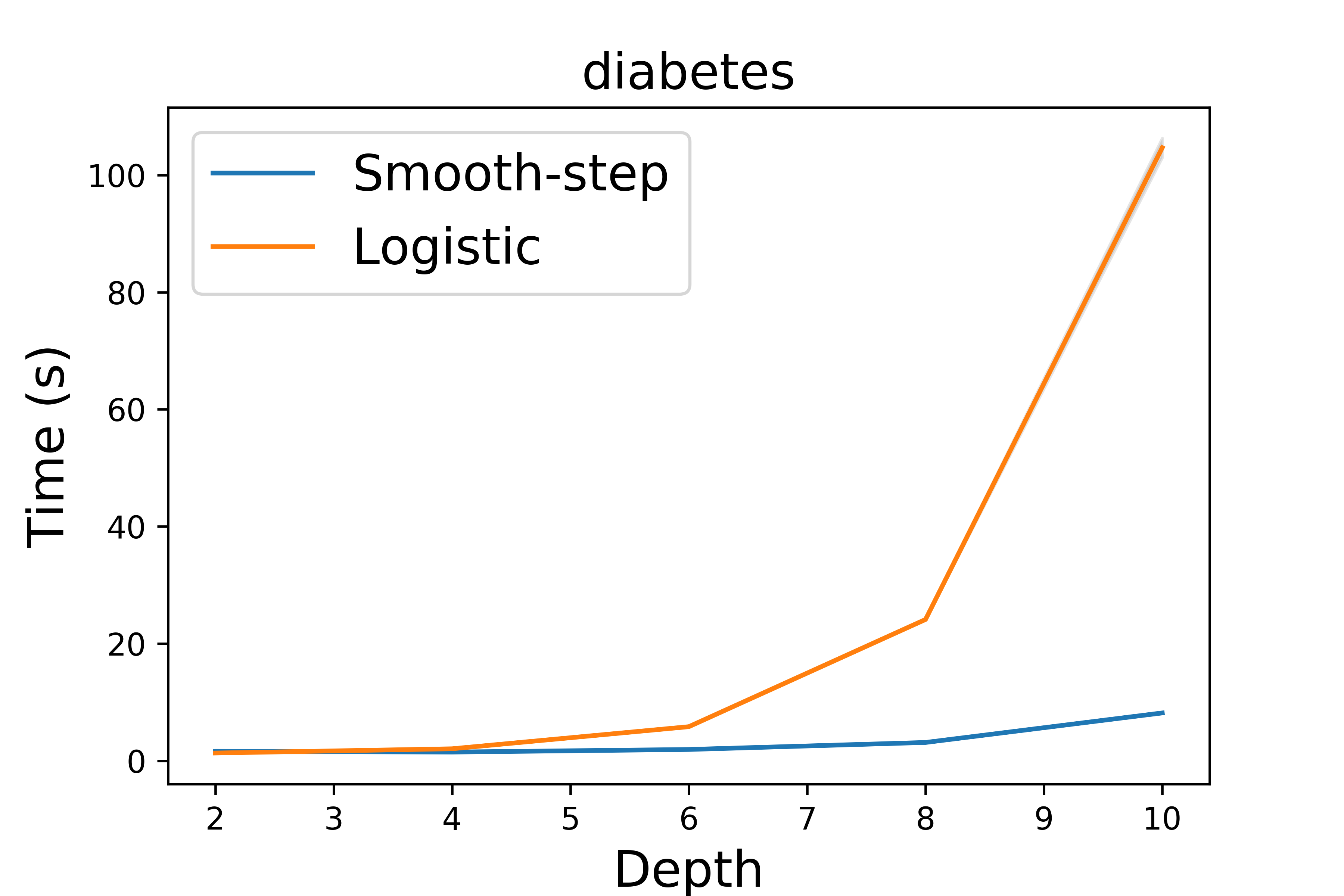} }}%
    \hspace{-0.5cm}
    \subfloat{{\includegraphics[width=5.8cm]{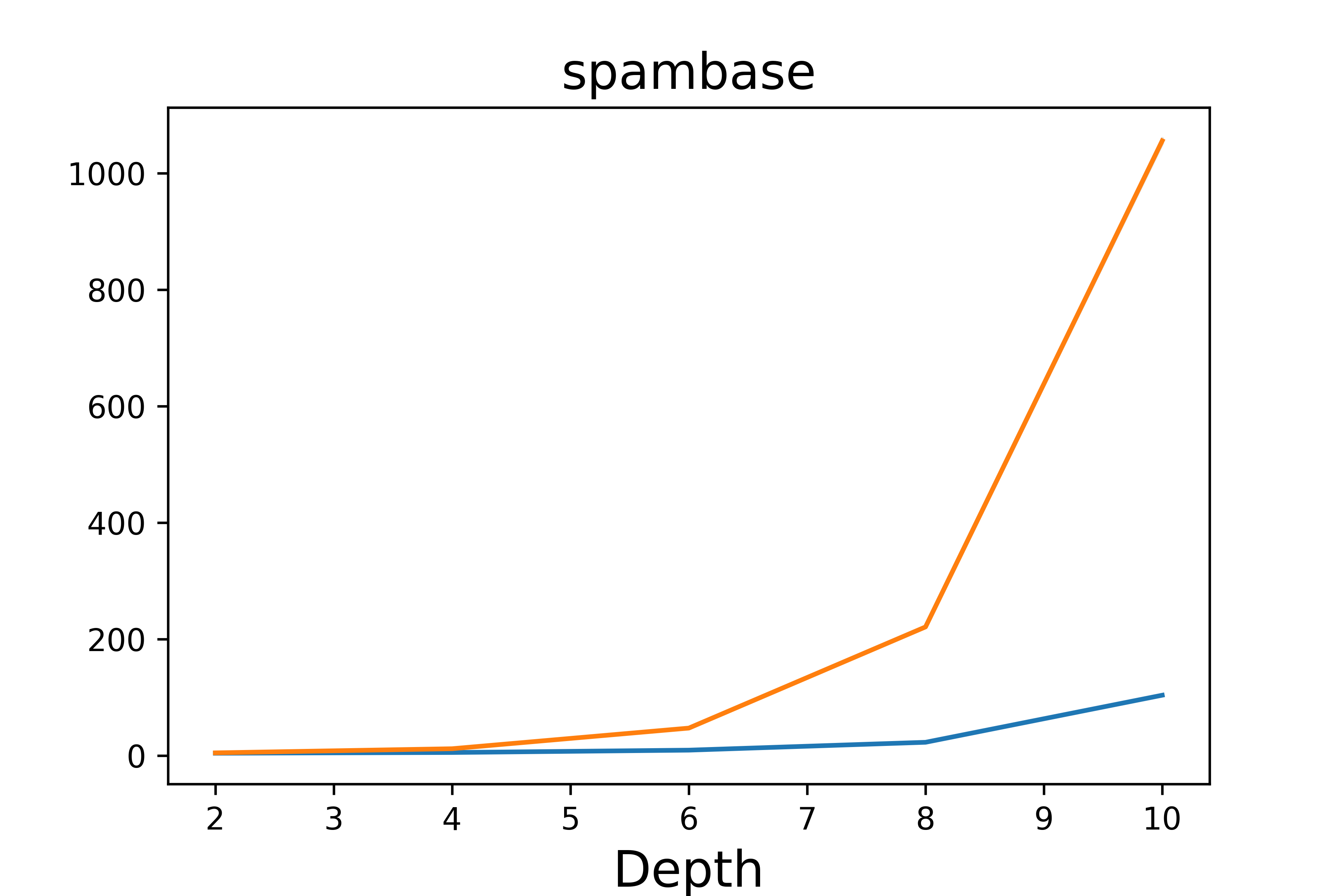} }}%
    \hspace{-0.5cm}
    \subfloat{{\includegraphics[width=5.8cm]{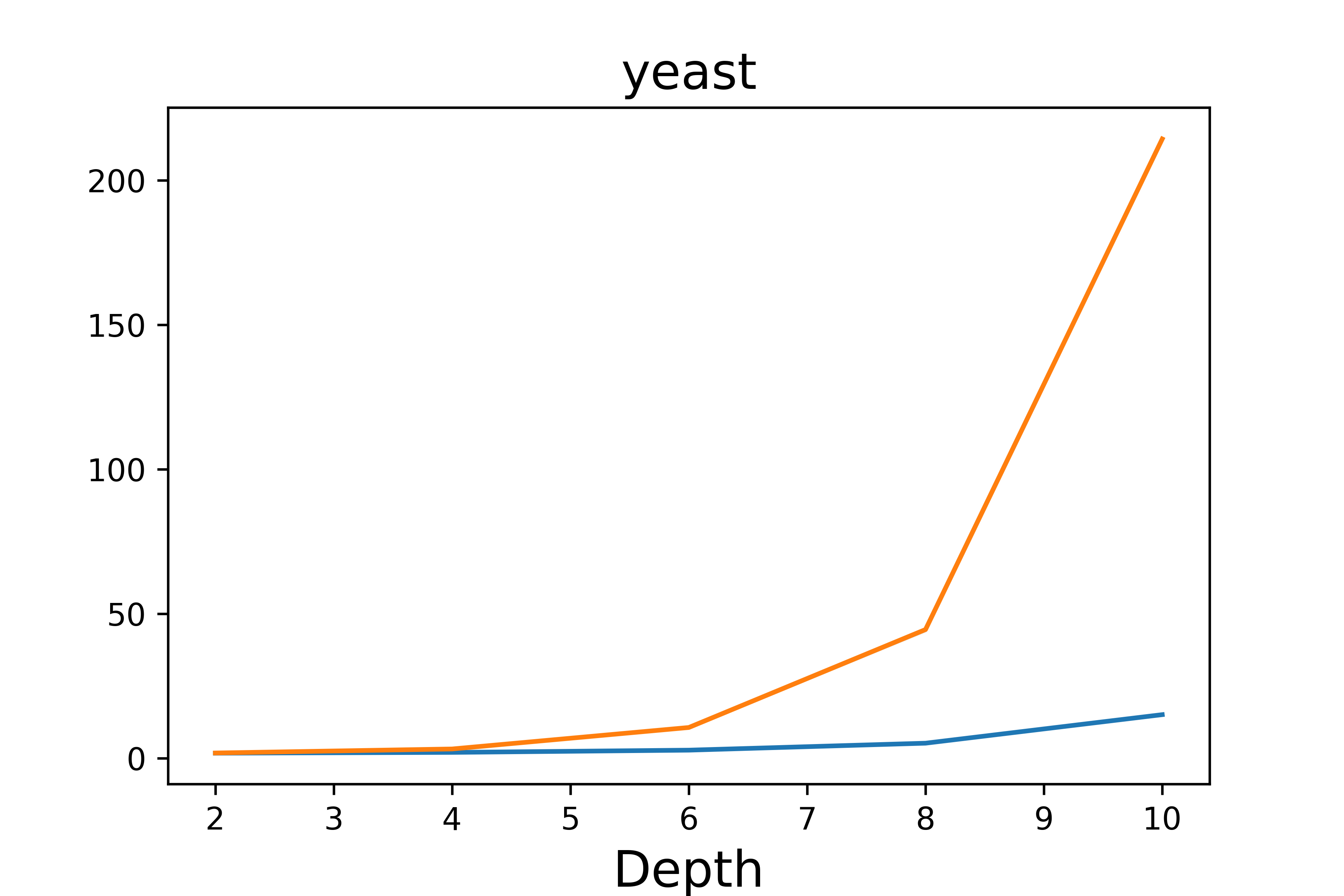} }}%
    \caption{Training time (sec) vs. tree depth for the smooth-step and logistic functions, averaged over 5 repetitions.}%
    \label{figure:time}
\end{figure*}

\textbf{Training Time: }
We measure the training time over $50$ epochs as a function of tree depth for both activation functions. We keep the same ensemble size ($10$) and use $\gamma = 1$ for the smooth-step as this corresponds to the worst-case training time (in the tuning range $[10^{-4}, 1]$), and we fix the optimization hyperparameters (batch size = 256 and learning rate = 0.1). We report the results for three of the datasets in Figure \ref{figure:time}; the results for the other datasets have very similar trends and are omitted due to space constraints. The results indicate a steep exponential increase in training time for the logistic activation after depth $6$. In contrast, the smooth-step has a slow growth, achieving over $10$x speed-up at depth $10$.
\subsection{TEL vs. Gradient Boosted Decision Trees} \label{sec:telvsxgb}
\paragraph{Predictive Performance: } We compare the predictive performance of TEL and GBDT on the 23 PMLB datasets, and we include L2-regularized logistic regression (LR) and CART as baselines. For a fair comparison, we use TEL as a standalone layer. For TEL and GBDT, we tune over the \# of trees, depth, learning rate, and L2 regularization. For TEL we also tune over the batch size, epochs, and $\gamma \in [10^{-4},1]$. For LR and CART, we tune the L2 regularization and depth, respectively. We use $50$ tuning rounds in Hyperopt with AUC as the  metric. We repeat the tuning/testing procedures on 15 random training/testing splits. The results are in Table \ref{tab:auc}.

As expected, no algorithm dominates on all the datasets. TEL outperforms GBDT on 9 datasets (5 are statistically significant). GBDT outperforms TEL on 8 datasets (7 of which are statistically significant). There were ties on the 6 remaining datasets; these typically correspond to easy tasks where an AUC of (almost) 1 can be attained. LR outperforms both TEL and GBDT on only 3 datasets with very marginal difference. Overall, the results indicate that TEL's performance is competitive with GBDT. Moreover, adding feature representation layers before TEL can potentially improve its performance further, e.g., see Section   \ref{sec:telvdense}.%, especially in image (convolutional layers), text (embeddings) and speech recognition.

\begin{table*}[htbp]
\centering
\caption{Test AUC on 23 PMLB datasets. Averages over 15 random repetitions are reported along with the SE. A star ($\bm{*}$) indicates statistical significance based on a paired two-sided t-test at a significance level of 0.05. Best results are in \textbf{bold}. }
\label{tab:auc}
\footnotesize
\begin{tabular}{@{}lllll@{}}
\toprule
Dataset                 & TEL               & GBDT              & L2 Logistic Reg.      & CART              \\ \midrule
ann-thyroid & $0.996 \pm 0.0$ & $\bm{1.0^{*}} \pm 0.0$ & $0.92 \pm 0.002$ & $0.997 \pm 0.0$ \\
breast-cancer-wisconsin & $\bm{0.995^{*}} \pm 0.001$ & $0.992 \pm 0.001$ & $0.991 \pm 0.001$ & $0.929 \pm 0.004$ \\
car-evaluation & $\bm{1.0} \pm 0.0$ & $\bm{1.0} \pm 0.0$ & $0.985 \pm 0.001$ & $0.981 \pm 0.001$ \\
churn & $0.916 \pm 0.004$ & $\bm{0.92^{*}} \pm 0.004$ & $0.814 \pm 0.003$ & $0.885 \pm 0.004$ \\
crx & $0.911 \pm 0.005$ & $\bm{0.933^{*}} \pm 0.004$ & $0.916 \pm 0.005$ & $0.905 \pm 0.005$ \\
dermatology & $\bm{0.998} \pm 0.001$ & $\bm{0.998} \pm 0.001$ & $\bm{0.998} \pm 0.001$ & $0.962 \pm 0.005$ \\
diabetes & $\bm{0.831^{*}} \pm 0.006$ & $0.82 \pm 0.006$ & $0.824 \pm 0.008$ & $0.774 \pm 0.008$ \\
dna & $0.993 \pm 0.0$ & $\bm{0.994^{*}} \pm 0.0$ & $0.991 \pm 0.0$ & $0.964 \pm 0.001$ \\
ecoli & $0.97^{*} \pm 0.003$ & $0.962 \pm 0.003$ & $\bm{0.972} \pm 0.003$ & $0.902 \pm 0.007$ \\
flare & $0.732 \pm 0.009$ & $\bm{0.738} \pm 0.01$ & $0.736 \pm 0.009$ & $0.717 \pm 0.01$ \\
heart-c & $0.903 \pm 0.006$ & $0.893 \pm 0.008$ & $\bm{0.908} \pm 0.005$ & $0.829 \pm 0.012$ \\
hypothyroid & $0.971 \pm 0.003$ & $\bm{0.987^{*}} \pm 0.002$ & $0.93 \pm 0.005$ & $0.926 \pm 0.011$ \\
nursery & $\bm{1.0} \pm 0.0$ & $\bm{1.0} \pm 0.0$ & $0.916 \pm 0.001$ & $0.996 \pm 0.0$ \\
optdigits & $\bm{1.0} \pm 0.0$ & $\bm{1.0} \pm 0.0$ & $0.998 \pm 0.0$ & $0.958 \pm 0.001$ \\
pima & $0.831 \pm 0.008$ & $0.825 \pm 0.006$ & $\bm{0.832} \pm 0.008$ & $0.758 \pm 0.011$ \\
satimage & $\bm{0.99} \pm 0.0$ & $\bm{0.99} \pm 0.0$ & $0.955 \pm 0.001$ & $0.949 \pm 0.001$ \\
sleep & $0.925 \pm 0.0$ & $\bm{0.927^{*}} \pm 0.0$ & $0.889 \pm 0.0$ & $0.876 \pm 0.001$ \\
solar-flare\_2 & $\bm{0.925} \pm 0.002$ & $0.924 \pm 0.002$ & $0.92 \pm 0.002$ & $0.907 \pm 0.002$ \\
spambase & $0.986 \pm 0.001$ & $\bm{0.989^{*}} \pm 0.001$ & $0.972 \pm 0.001$ & $0.926 \pm 0.002$ \\
texture & $\bm{1.0} \pm 0.0$ & $\bm{1.0} \pm 0.0$ & $\bm{1.0} \pm 0.0$ & $0.974 \pm 0.001$ \\
twonorm & $\bm{0.998^{*}} \pm 0.0$ & $0.997 \pm 0.0$ & $\bm{0.998} \pm 0.0$ & $0.865 \pm 0.002$ \\
vehicle & $\bm{0.953^{*}} \pm 0.003$ & $0.931 \pm 0.002$ & $0.941 \pm 0.002$ & $0.871 \pm 0.004$ \\
yeast & $\bm{0.861} \pm 0.004$ & $0.859 \pm 0.004$ & $0.852 \pm 0.004$ & $0.779 \pm 0.005$ \\
\midrule
\textit{\# wins} & \textit{12} & \textit{14} & \textit{6} & \textit{0} \\
\bottomrule
\end{tabular}
\end{table*}
%\vspace{-2cm}
\begin{figure*}[htbp]
    \centering
    \subfloat{{\includegraphics[width=5.8cm]{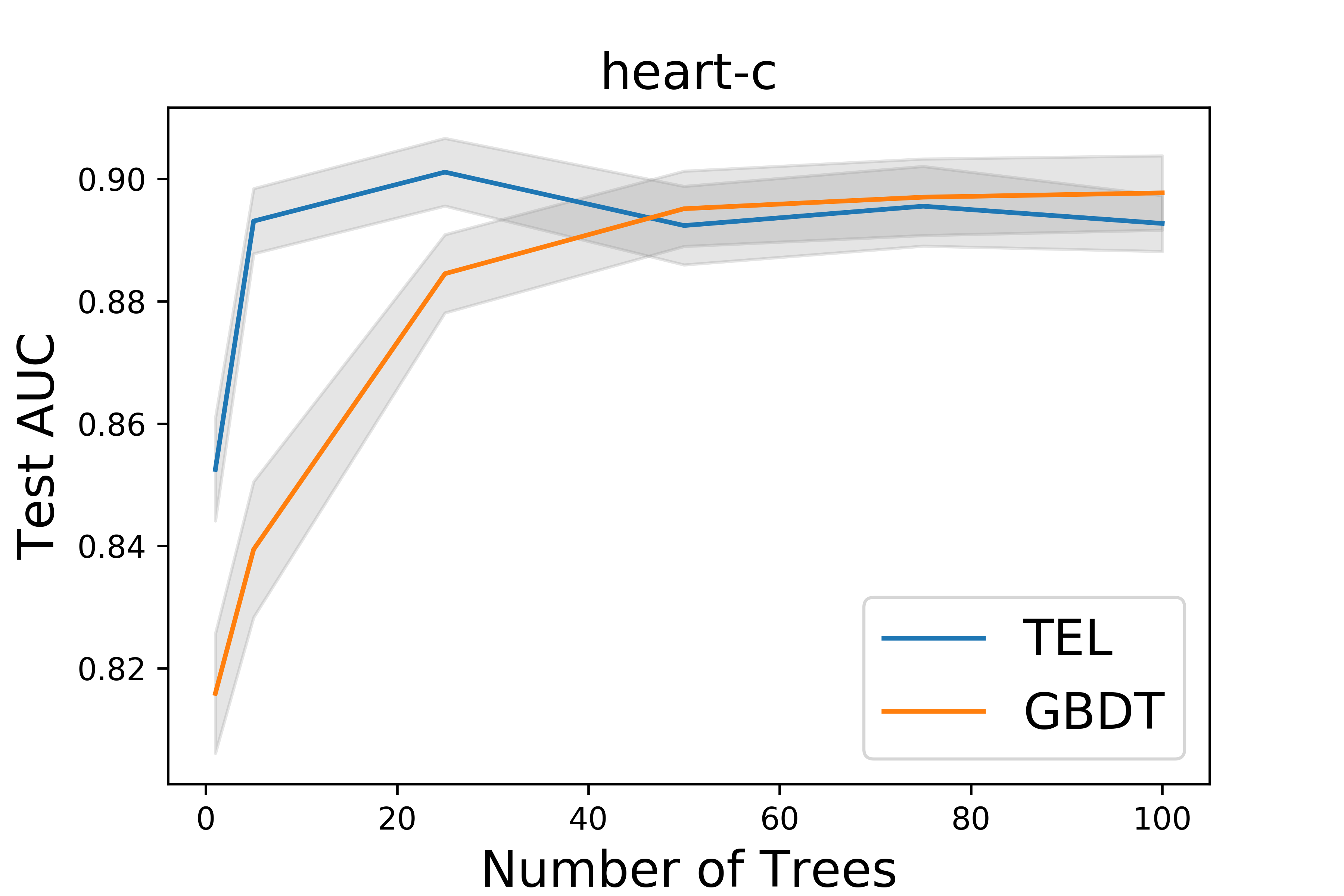} }}%
    \hspace{-0.2cm}
    \subfloat{{\includegraphics[trim={0.7cm 0cm 0cm 0cm},clip,width=5.5cm]{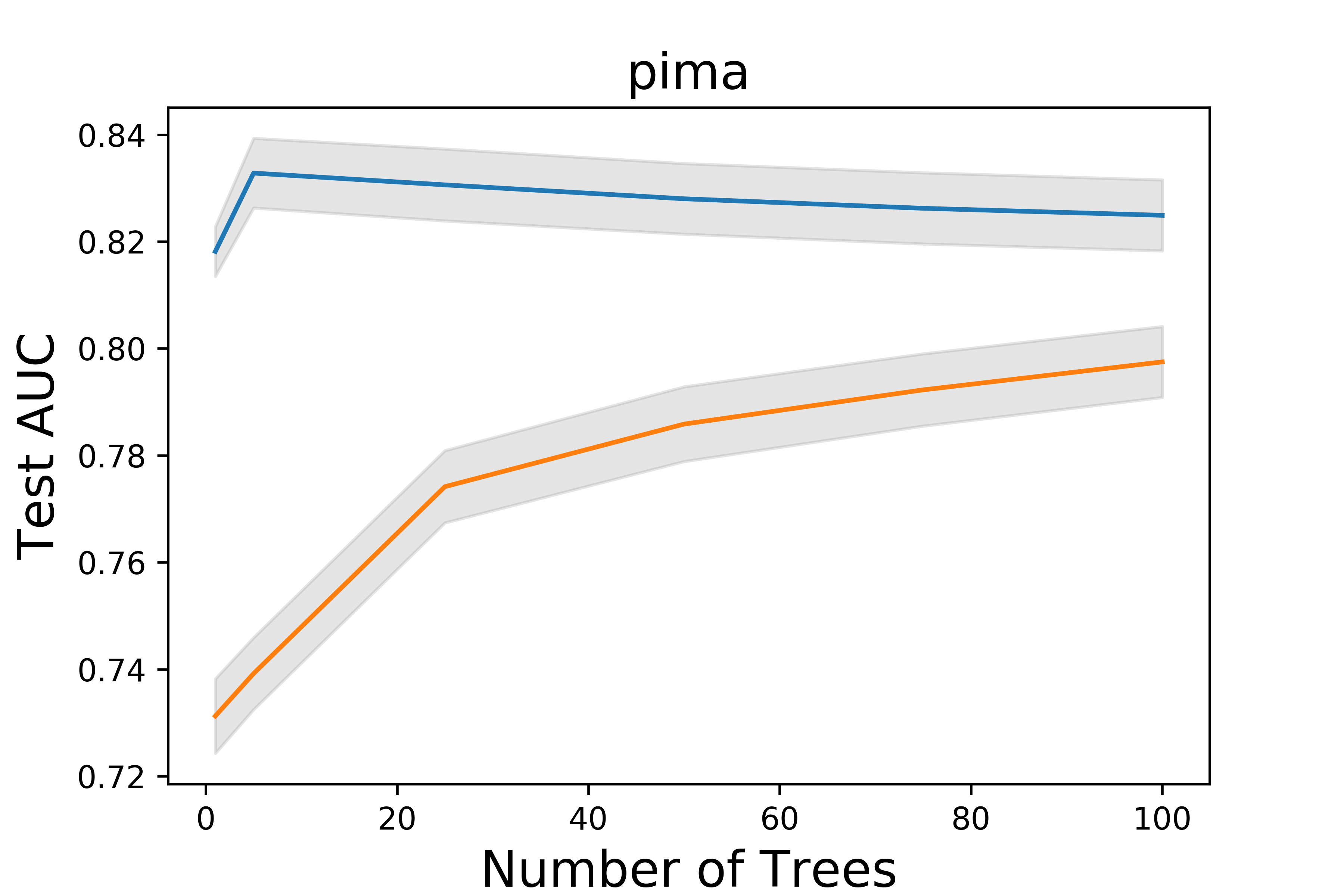} }}%
    \hspace{-0.5cm}
    \subfloat{{\includegraphics[width=5.8cm]{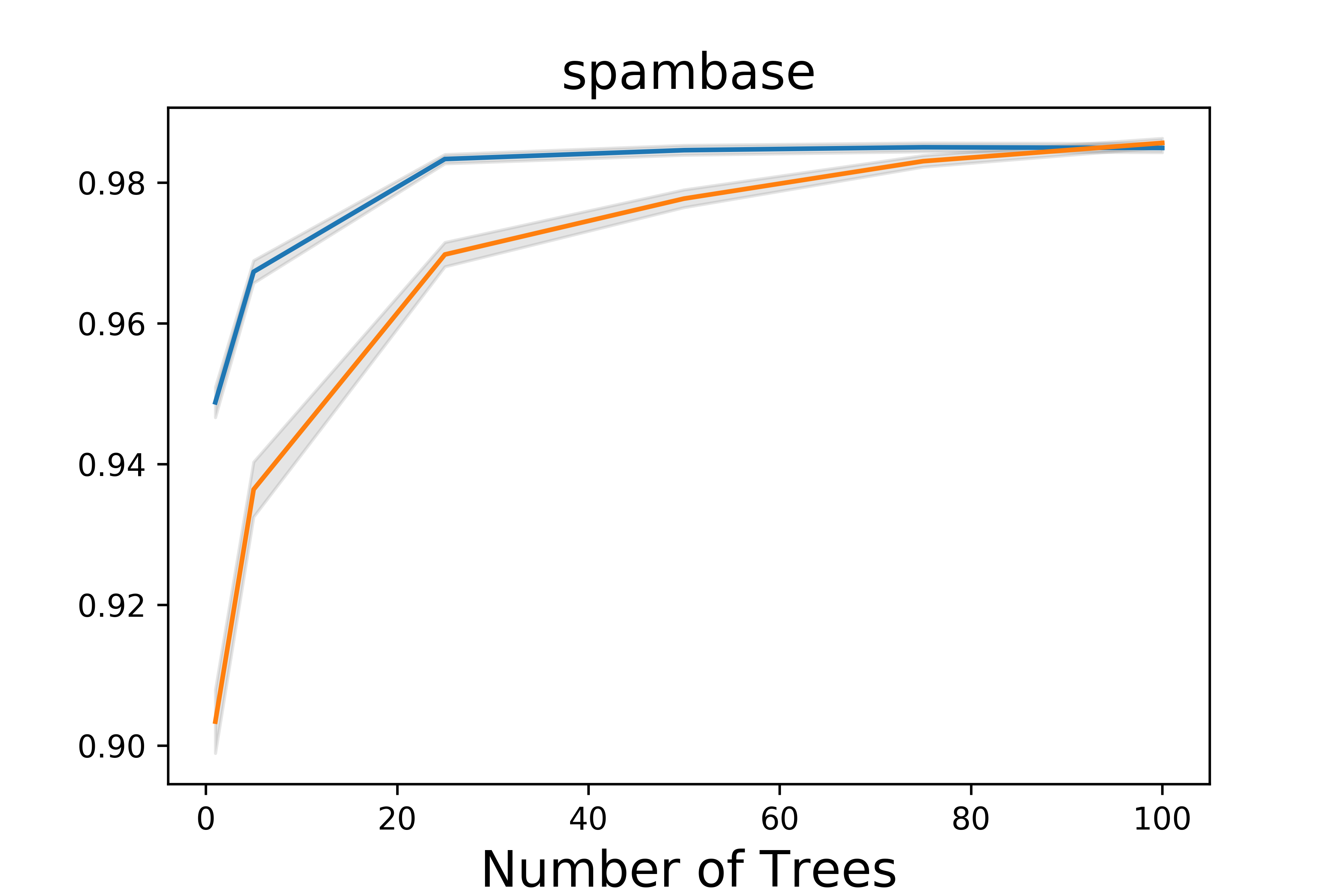} }}%
    \caption{Mean test AUC vs \# of trees (15 trials). SE is shaded. TEL and GBDT have (roughly) the same \# of params/tree. }%
    \label{fig:aucvsnumtrees}
\end{figure*}

\begin{table*}[htbp]
\centering
\caption{Average and SE for test accuracy, loss and \# of params for CNN-Dense and CNN-TEL over $5$ random initializations. A star $\bm{*}$ indicates statistical significance based on a paired two-sided t-test at a level of $5\%$. Best values are in \textbf{bold}.}
\footnotesize
\label{table:densevtel}
\begin{tabular}{@{}lcccccc@{}}
\toprule
              & \multicolumn{3}{c}{CNN-Dense}                                                 & \multicolumn{3}{c}{CNN-TEL}                              \\ \midrule
Dataset       & Accuracy            & Loss                  & \multicolumn{1}{c|}{\# Params} & Accuracy            & Loss                  & \# Params \\
CIFAR10       & $0.7278 \pm 0.0047$ & $1.673 \pm 0.170$     & \multicolumn{1}{c|}{$7,548,362$}  & $\bm{0.7296} \pm 0.0109$ & $\bm{1.202^{*}} \pm 0.011$     & $\bm{926,465}$    \\
MNIST         & $0.9926 \pm 0.0002$ & $0.03620 \pm 0.00121$ & \multicolumn{1}{c|}{$5,830,538$}  & $\bm{0.9930} \pm 9e-5$   & $\bm{0.03379} \pm 0.00093$ & $\bm{699,585}$    \\
Fashion MNIST & $\bm{0.9299} \pm 0.0012$ & $0.6930 \pm 0.0291$   & \multicolumn{1}{c|}{$5,567,882$}  & $0.9297 \pm 0.0012$ & $\bm{0.3247^{*}} \pm 0.0045$   & $\bm{699,585}$    \\ \bottomrule
\end{tabular}
\end{table*}

\textbf{Compactness and Sensitivity: } We compare the number of trees and sensitivity of TEL and GBDT on datasets from Table \ref{tab:auc} where both models achieve comparable AUCs---namely, the heart-c, pima and spambase datasets. With similar predictive performance,  compactness can be an important factor in choosing a model over the other. For TEL, we use the models trained in Table \ref{tab:auc}. As for GBDT, for each dataset, we fix the depth so that the number of parameters per tree in GBDT (roughly) matches that of TEL. We tune over the main parameters of GBDT (50 iterations of Hyperopt, under the same parameter ranges of Table \ref{tab:auc}). We plot the test AUC versus the number of trees in Figure \ref{fig:aucvsnumtrees}. On all datasets, the test AUC of TEL peaks at a significantly smaller number of trees compared to GBDT. For example, on pima, TEL's AUC peaks at 5 trees, whereas GBDT requires more than 100 trees to achieve a comparable performance---this is more than $20$x reduction in the number of parameters. Moreover, the performance of TEL is less sensitive w.r.t. to changes in the number of trees. These observations can be attributed to the joint optimization performed in TEL, which can lead to more expressive ensembles compared to the stage-wise optimization in GBDT. 

\subsection{TEL vs. Dense Layers in CNNs} \label{sec:telvdense}
We study the potential benefits of replacing dense layers with TEL in CNNs, on the CIFAR-10, MNIST, and Fashion MNIST datasets. We consider 2 convolutional layers, followed by intermediate layers (max pooling, dropout, batch normalization), and finally dense layers; we refer to this as CNN-Dense. We also consider a similar architecture, where the final dense layers are replaced with a single dense layer followed by TEL; we refer to this model as CNN-TEL. We tune over the optimization hyperparameters, the number of filters in the convolutional layers, the number and width of the dense layers, and the different parameters of TEL (see appendix for details). We run Hyperopt for 25 iterations with classification accuracy as the target metric. After tuning, the models are trained using 5 random weight initializations. 

The classification accuracy and loss on the test set and the total number of parameters are reported in Table \ref{table:densevtel}. While the accuracies are comparable, CNN-TEL achieves a lower test loss on the three datasets, where the $28\%$ and $53\%$ relative improvements on CIFAR and Fashion MNIST are statistically significant. Since we are using cross-entropy loss, this means that TEL gives higher scores on average, when it makes correct predictions. Moreover, the number of parameters in CNN-TEL is $\sim 8$x smaller than CNN-Dense. This example also demonstrates how representation layers can be effectively leveraged by TEL---GBDT's performance is significantly lower on MNIST and CIFAR-10, e.g., see the comparisons in \citet{ponomareva2017compact}.
\vspace{-0.1cm}
\textcolor{black}{\section{Conclusion and Future Work}}
We introduced the tree ensemble layer (TEL) for neural networks. The layer is composed of an additive model of differentiable decision trees that can be trained end-to-end with the neural network, using first-order methods. Unlike differentiable trees in the literature, TEL supports conditional computation, i.e., each sample is routed through a small part of the tree's architecture. This is achieved by using the smooth-step activation function for routing samples, along with specialized forward and backward passes for reducing the computational complexity. Our experiments indicate that TEL achieves competitive predictive performance compared to gradient boosted decision trees (GBDT) and dense layers, while leading to significantly more compact models. In addition, by effectively leveraging convolutional layers, TEL significantly outperforms GBDT on multiple image classification datasets. 

One interesting direction for future work is to equip TEL with mechanisms for exploiting feature sparsity, which can further speed up computation. Promising works in this direction include feature bundling \citep{ke2017lightgbm} and learning under hierarchical sparsity assumptions \citep{hazimeh2020learning}. Moreover, it would be interesting to study whether the smooth-step function, along with specialized optimization methods, can be an effective alternative to the logistic function in other machine learning models. 

% One interesting direction for future work is to equip TEL with principled  mechanisms for pruning and handling categorical features, which have the potential to further improve performance. Moreover, it would be interesting to study whether the smooth-step function, along with specialized optimization  methods, can be an effective alternative to the logistic function in other machine learning models. 

\clearpage

\section*{Acknowledgements} We would like to thank Mehryar Mohri for the useful discussions. Part of this work was done when Hussein Hazimeh was at Google Research. At MIT, Hussein acknowledges research funding from the Office of Naval Research (ONR-N000141812298). Rahul Mazumder acknowledges research funding from the Office of Naval Research (ONR-N000141812298, Young Investigator Award), the National Science Foundation (NSF-IIS-1718258), and IBM.
\bibliography{references}
\bibliographystyle{icml2020}

\appendix
\setcounter{table}{0}
\renewcommand{\thetable}{\thesection.\arabic{table}}
\setcounter{figure}{0}
\renewcommand\thefigure{\thesection.\arabic{figure}}
\section{Notation}
Table \ref{tab:notations} lists the notation used throughout the paper.

\begin{table*}[htbp]
    \renewcommand{\arraystretch}{1.1}
    \caption{List of notation used.}
    \footnotesize
    \label{tab:notations}
    \centering
    \begin{tabular}{|c|c|p{120mm}|}
    \hline
    \textbf{Notation}     & \textbf{Space or Type} & \textbf{Explanation}  \\
    \hline
    $\mathcal{X}$ &$\mathbb{R}^{p}$ & Input feature space.\\
    \hline
    $\mathcal{Y}$ & $\mathbb{R}^{k}$ & Output (label) space. \\
    \hline
    $m$ & $\mathbb{Z}_{> 0}$ & Number of trees in the TEL. \\
    \hline
    $\mathcal{T}(x)$ & Function & The output of TEL, a function that takes an input sample and returns a logit which corresponds to the sum of all the trees in the ensemble. Formally, $\mathcal{T}: \mathcal{X} \to \mathbb{R}^{k}.$ \\
    \hline
    $T(x)$ & Function & A single perfect binary tree which takes an input sample and returns a logit, i.e., $T: \mathcal{X} \to \mathbb{R}^{k}$.\\
    \hline
    $d$ & $\mathbb{Z}_{> 0}$ & The depth of tree $T$. \\
    \hline
    $\mathcal{I}$ & Set & The set of internal (split) nodes in $T$. \\
    \hline
    $\mathcal{L}$ & Set & The set of leaf nodes in $T$. \\
    \hline
    $\mathcal{A}(i)$ & Set & The set of ancestors of node $i$.\\ 
    \hline
    $\{ x \to i \}$ & Event & The event that sample  $x \in \mathbb{R}^{p}$ reaches node $i$. \\
    \hline
    $w_i$ & $\mathbb{R}^{p}$ & Weight vector of internal node $i$ (trainable). Defines the hyperplane split used in sample routing. \\
    \hline
    $W$ & $\mathbb{R}^{|\mathcal{I}| \times p}$ & Matrix of all the internal nodes weights. \\
    \hline
    $\mathcal{S}$ & Function & Activation function $\mathbb{R} \to [0,1]$ \\
    \hline 
    $\mathcal{S}(\langle w_i, x \rangle)$ & $[0,1]$ & Probability (proportion) that internal node $i$ routes $x$ to the left. \\
    \hline
    $[l \swarrow i]$ & Event & The event that leaf $l$ belongs to the left subtree of node $i \in \mathcal{I}$. \\
    \hline
    $[l \searrow i]$ & Event & The event that leaf $l$ belongs to the right subtree of node $i \in \mathcal{I}$. \\
    \hline
     $o_{l}$ & $\mathbb{R}^{k}$ & Leaf $l$'s weight vector (trainable). \\
    \hline
    $O$ & $\mathbb{R}^{|\mathcal{L}| \times k}$ & Matrix of leaf weights. \\
    \hline
    $\gamma$ & ${\mathbb{R}}_{\geq 0}$ & Non-negative scaling parameter for the smooth-step activation function. \\
    \hline
    $L$ & Function & Loss function for training (e.g., cross-entropy). \\ 
    \hline
    $U$, $N$ & $\mathbb{Z}_{>0}$ & Number of leaves and internal nodes, respectively, that a sample $x$ reaches. \\ 
    \hline
    $\mathcal{R}$ & Set &  The set of reachable leaves. \\
    \hline 
    $\mathcal{F}$ & Set & The set of ancestors of the reachable leaves, whose activation is fractional, i.e., $\mathcal{F} = {\{i \in \mathcal{I} \ | \  i \in \mathcal{A}(l), \  l \in \mathcal{R}, \  0 < \mathcal{S}(\langle x, w_i \rangle) < 1  \}}$. \\
    \hline
    \end{tabular}
\end{table*}

\section{Appendix for Section \ref{section:tree}}
Figure \ref{fig:reachable_leaves} was generated by training a single tree with depth $10$ using the smooth-step activation function. We optimized the cross-entropy loss using Adam \cite{kingma2014adam} with base learning rate $=0.1$ and batch size $=256$. The y-axis of the plot corresponds to the average number of reachable leaves per sample (each point in the graph corresponds to a batch, and averaging is done over the samples in the batch). For details on the diabetes dataset used in this experiment and the computing setup, please refer to Section \ref{sec:appendixexp} of the appendix.

\section{Appendix for Section \ref{sec:conditional}}

\subsection{Example of Conditional Forward and Backward Passes}
\begin{figure*}[htbp]
    \centering
    \subfloat{{\includegraphics[width=5.5cm, height=5.5cm]{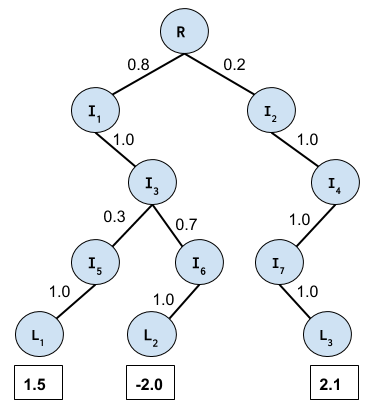}}}%
    \hspace{2cm}
    \subfloat{\raisebox{6em}{\includegraphics[trim={1cm 4cm 5cm 2cm},clip, width=3.2cm]{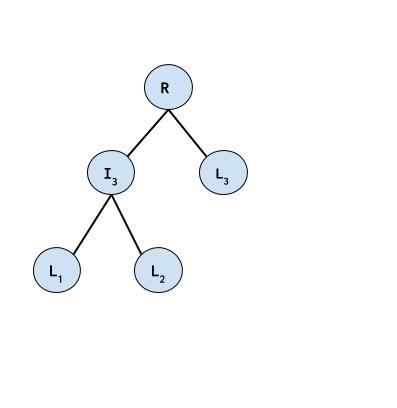} }}%
    \caption{\textbf{Left}: Reachable sub-tree for the conditional forward pass. \textbf{Right}: Corresponding (fractional) tree for the conditional backward pass where most of the internal (splitting) nodes of the forward pass sub-tree have been eliminated.}%
    \label{figure:tel_fb}
\end{figure*}

Figure~\ref{figure:tel_fb} shows an example of the tree traversed by the conditional forward pass, along with the corresponding $T_{fractional}$ used during the backward pass,  for a simple regression tree of depth $d=4.$ Note that only 3 leaves are reachable (out of the 16 leaves of a perfect, depth-4 tree). The values inside the boxes at the bottom correspond to the regression values (scalars) stored at each leaf. The output of the tree for the forward pass shown on the left of Figure~\ref{figure:tel_fb}, is given by
\[
T(x) = 0.8 \cdot 0.3 \cdot 1.5 + 0.8 \cdot 0.7 \cdot (-2.0) + 0.2 \cdot 2.1 = -0.34.
\]
Note also that the tree used to compute the backward pass is substantially smaller than the forward pass tree since all the hard-routing internal nodes of the latter have been removed.

\subsection{Memory Complexity of the Conditional Forward Pass} The memory requirements depend on whether the forward pass is being used for inference or training. For inference, a node can be discarded as soon as it is traversed. Since we traverse the tree in a depth-first manner, the worst-case memory complexity is $\mathcal{O}(d)$. When used in the context of training, additional quantities need to be stored in order to perform the backward pass efficiently. In particular, we need to store $l.prob$ for every reachable leaf $l$, and $\langle w_i, x \rangle$ for every internal node $i \in \mathcal{F}$, where $\mathcal{F}$ is set of ancestors of the reachable leaves whose activation is fractional---see Section \ref{sec:backward} for a formal definition of $\mathcal{F}$. Note that $|F| = U - 1$ (as discussed after Definition \ref{def:fractional}). Thus, the worst-case memory complexity when used in the context of training is $\mathcal{O}(d + U)$.

\subsection{Time Complexity of the Conditional Backward Pass} 
Lines 8 and 9 perform $\mathcal{O}(k)$ operations so each leaf requires $\mathcal{O}(k)$ operations. Lines 11 and 12 are $\mathcal{O}(1)$ since for every $i \in \mathcal{F}$, $\langle w_i, x \rangle$ is available from the conditional forward pass. Lines 13 and 14 are $\mathcal{O}(p)$, while line 15 is $\mathcal{O}(1)$. The total number of internal nodes traversed is $|\mathcal{F}|$. Moreover, we always have $|\mathcal{F}| = U - 1$ (see the discussion following Definition \ref{def:fractional}). Therefore, the worst-case complexity is $\mathcal{O}(Up + Uk)$. By Theorem 1, the maximum number of non-zero entries in the three gradients is $p + p|\mathcal{F}| + Uk = \mathcal{O}(Up + Uk)$ (and it is easy to see that this rate is achievable). The best-case complexity of Algorithm 2 is $\mathcal{O}(k)$---this corresponds to the case where there is only one reachable leaf ($U=1$), so the fractional tree is composed of a single node.

\subsection{Proof of Theorem 1}
\paragraph{Gradient of Loss w.r.t. $x$: }
By the chain rule, we have: 
\begin{align} \label{eq:gradLxi}
\underbrace{\fracpartial{L}{x}}_{1 \times p}  = {\underbrace{\fracpartial{L}{T}}_{1 \times k} \underbrace{\fracpartial{T}{x}}_{k \times p}}
\end{align}
The first term in the RHS above is available from backpropagation. The second term can be written more explicitly as follows:
\begin{align*}
\fracpartial{T}{x} & = \fracpartial{\sum_{l \in \mathcal{L}} P(\{ x \to l \}) o_{l}}{x} \\ 
& = \sum_{l \in \mathcal{L}} o_{l} \frac{\partial}{\partial x} \prod_{j \in \mathcal{A}(l)} r_{j,l}(x) \\
& = {\sum_{l \in \mathcal{L}} o_{l} \sum_{i \in \mathcal{A}(l)}  \frac{\partial}{\partial x} r_{i,l}(x) \prod_{j \in \mathcal{A}(l), j \neq i} r_{j,l}(x)}
\end{align*}
We make three observations that allows us to simplify the expression above. 
First, if a leaf $l$ is not reachable by $x$, then the inner term in the second summation above must be $0$. This means that the outer summation can be restricted to the set of reachable leaves  $\mathcal{R}$. Second, if an internal node $i$ has a non-fractional $r_{i,l}(x)$, then  $\frac{\partial}{\partial x} r_{i,l}(x) = 0$. This implies that we can restrict the inner summation to be only over  $\mathcal{A}(l) \cap \mathcal{F}$. Third, the second term in the inner summation can be simplified by noting that the following holds for any $l \in \mathcal{R}$: 
\begin{align} \label{eq:prodcutsimplification}
\prod_{j \in \mathcal{A}(l), j \neq i} r_{j,l}(x) = \frac{P(\{ x \to l \})}{r_{i,l}(x)}
\end{align}
Note that in the above, $r_{i,l}(x)$ cannot be zero since $l \in \mathcal{R}$ (otherwise, $l$ will be unreachable). Combining the three observations above, $\fracpartial{T}{x}$ simplifies to:
\begin{align} \label{eq:delT_delx_simplified}
\fracpartial{T}{x} & = {\sum_{l \in \mathcal{R}} o_{l} \sum_{i \in \mathcal{A}(l) \cap \mathcal{F}}  \frac{\partial}{\partial x} r_{i,l}(x) \frac{P(\{ x \to l \})}{r_{i,l}(x)}}.
\end{align}
Note that: 
\begin{align} \label{eq:derivativeS}
& \frac{\partial}{\partial x} r_{i,l}(x) =  \mathcal{S}'(\langle x, w_i \rangle) w_i^T (-1)^{\mathds{1}[i \searrow l]}.
\end{align}
Plugging \eqref{eq:derivativeS} into \eqref{eq:delT_delx_simplified},  and then using \eqref{eq:gradLxi}, we get:
\begin{align} \label{eq:gradlx}
\fracpartial{L}{x} = \sum_{l \in \mathcal{R}} g(l) \sum_{i \in \mathcal{A}(l) \cap \mathcal{F}}  w_i^T  (-1)^{\mathds{1}[i \searrow l]} \frac{\mathcal{S}'(\langle x, w_i \rangle)}{r_{i,l}(x)}
\end{align}
where 
\begin{align}   
g(l) & =  P(\{ x \to l \}) \langle \fracpartial{L}{T},  o_{l} \rangle. \label{eq:g}
\end{align}
Finally, we switch the order of the two summations in \eqref{eq:gradlx} to get:
\begin{align*}
\fracpartial{L}{x} & = \sum_{i \in \mathcal{F}} \sum_{l \in \mathcal{R} | i \in \mathcal{A}(l)} g(l) w_i^T  (-1)^{\mathds{1}[i \searrow l]} \frac{\mathcal{S}'(\langle x, w_i \rangle)}{r_{i,l}(x)} \\
& = \sum_{i \in \mathcal{F}}  \frac{\mathcal{S}'(\langle x, w_i \rangle)}{\mathcal{S}(\langle x, w_i \rangle)} w_i^T \sum_{l \in \mathcal{R} | [l \swarrow i]} g(l) \\
& - \sum_{i \in \mathcal{F}}  \frac{\mathcal{S}'(\langle x, w_i \rangle)}{1 - \mathcal{S}(\langle x, w_i \rangle)} w_i^T \sum_{l \in \mathcal{R} | [i \searrow l]} g(l)
\end{align*}
\paragraph{Gradient of Loss w.r.t. $w_i$: } By the chain rule: 
\begin{align}
\fracpartial{L}{w_i} =\fracpartial{L}{T} \fracpartial{T}{w_i}
\end{align}
The first term in the summation is provided from backpropagation. The second term can be simplified as follows:
\begin{align} 
\fracpartial{T}{w_i} & = \fracpartial{\sum_{l \in \mathcal{L}} P(\{ x \to l \}) o_{l}}{w_i} \nonumber \\ 
& = \sum_{l \in \mathcal{L}} o_{l} \frac{\partial}{\partial w_i} \prod_{j \in \mathcal{A}(l)} r_{j,l}(x) \nonumber \\
& = \sum_{l \in \mathcal{L} | i \in \mathcal{A}(l)} o_{l} \frac{\partial}{\partial w_i} r_{i,l}(x) \prod_{j \in \mathcal{A}(l), j \neq i} r_{j,l}(x), \label{eq:gradTiwz}
\end{align}
If $i \in \mathcal{F}^c$ then the term inside the summation above must be zero, which leads to $\fracpartial{T}{w_i} = 0$.

Next, we assume that $i \in \mathcal{F}$. If if a leaf $l$ is not reachable, then the term inside the summation of \eqref{eq:gradTiwz} is zero. Using this observation along with \eqref{eq:prodcutsimplification}, and simplifying we get the following for every $i \in \mathcal{F}$:
\begin{align*}
\fracpartial{L}{w_i} & =  \frac{\mathcal{S}'(\langle x, w_i \rangle)}{\mathcal{S}(\langle x, w_i \rangle)} x^T \sum_{l \in \mathcal{R} | [l \swarrow i]} g(l) \\
& -  \frac{\mathcal{S}'(\langle x, w_i \rangle)}{1 - \mathcal{S}(\langle x, w_i \rangle)} x^T \sum_{l \in \mathcal{R} | [i \searrow l]} g(l)
\end{align*}

\paragraph{Gradient of Loss w.r.t. O: } Note that 
\begin{align}
\fracpartial{T}{o_l} & = \fracpartial{\sum_{v \in \mathcal{L}} P(\{ x \to v \}) o_{v}}{o_l} \\ 
& = P(\{ x \to l \}) I_k,
\end{align}
where $I_k$ is the $k \times k$ identity matrix. Applying the chain rule, we get:  
\begin{align} \label{eq:gradLO}
\fracpartial{L}{o_l} = \fracpartial{L}{T} P(\{ x \to l \}).
\end{align}

\section{Appendix for Section \ref{label-experiments}} \label{sec:appendixexp}

\textbf{Datasets: } We consider 23 classification datasets from the Penn Machine Learning Benchmarks (PMLB) repository\footnote{\url{https://github.com/EpistasisLab/penn-ml-benchmarks}} \cite{Olson2017PMLB}. No additional preprocessing was done as the PMLB datasets are already preprocessed---see  \citet{Olson2017PMLB} for details. We randomly split each of the PMLB datasets into 70\% training and 30\% testing sets. The three remaining datasets are CIFAR-10 \cite{krizhevsky2009learning}, MNIST \cite{lecun1998gradient}, and Fashion MNIST \cite{xiao2017fashion}. For these, we kept the original training/testing splits (60K/10K for MNIST and Fashion MNIST, and 50K/10K for CIFAR) and normalized the pixel values to the range $[0,1]$. A summary of the 26 datasets considered is in Table \ref{table:datasets}.

\begin{table}[htbp]
\centering
\caption{Dataset Statistics}
\footnotesize
\label{table:datasets}
\begin{tabular}{@{}lccc@{}}
\toprule
Dataset                 & \# samples     & \# features   & \# classes  \\ \midrule
ann-thyroid             & 7200   & 21  & 3  \\
breast-cancer-w. & 569    & 30  & 2  \\
car-evaluation          & 1728   & 21  & 4  \\
churn                   & 5000   & 20  & 2  \\
CIFAR                   & 60000  & 3072 & 10 \\
crx                     & 690    & 15  & 2  \\
dermatology             & 366    & 34  & 6  \\
diabetes                & 768    & 8   & 2  \\
dna                     & 3186   & 180 & 3  \\
ecoli                   & 327    & 7   & 5  \\
Fashion MNIST           & 70000  & 784 & 10 \\
flare                   & 1066   & 10  & 2  \\
heart-c                 & 303    & 13  & 2  \\
hypothyroid             & 3163   & 25  & 2  \\
MNIST                   & 70000  & 784 & 10 \\
nursery                 & 12958  & 8   & 4  \\
optdigits               & 5620   & 64  & 10 \\
pima                    & 768    & 8   & 2  \\
satimage                & 6435   & 36  & 6  \\
sleep                   & 105908 & 13  & 5  \\
solar-flare\_2           & 1066   & 12  & 6  \\
spambase                & 4601   & 57  & 2  \\
texture                 & 5500   & 40  & 11 \\
twonorm                 & 7400   & 20  & 2  \\
vehicle                 & 846    & 18  & 4  \\
yeast                   & 1479   & 8   & 9  \\ \bottomrule
\end{tabular}
\end{table}

\textbf{Computing Setup: } We used a cluster running CentOS 7 and  equipped with Intel Xeon Gold 6130 CPUs (with a 2.10GHz clock). The tuning and training was done in parallel over the competing models and datasets (i.e., each (model,dataset) pair corresponds to a separate job). For the experiments of Sections \ref{sec:softexperiment} and \ref{sec:telvsxgb}, each job involving TEL and XGBoost was restricted to 4 cores and 8GB of RAM, whereas LR and CART were restricted to 1 core and 2GB. The jobs in the experiment of Section \ref{sec:telvdense} were each restricted to 8 cores and 32GB RAM. We used Python 3.6.9 to run the experiments with the following libraries: TensorFlow 2.1.0-dev20200106, XGBoost 0.90, Sklearn 0.19.0, Hyperopt 0.2.2, Numpy 1.17.4, Scipy 1.4.1, and GCC 6.2.0 (for compiling the custom forward/backward passes). 

\subsection{Tuning Parameters and Architectures: } A list of all the tuning parameters and their distributions is given for every experiment below. For experiment \ref{sec:telvdense}, we also describe the architectures used in detail.

\textbf{Experiment of Section \ref{sec:softexperiment}}  For the predictive performance experiment, we use the following: 
\begin{itemize}
    \item Learning rate: Uniform over $\{10^{-1}$, $10^{-2}$, \dots, $10^{-5}\}$.
    \item Batch size: Uniform over $\{32, 64, 128, 256, 512\}$.
    \item Number of Epochs: Discrete uniform with range $[5,100]$.
    \item $\alpha$: Log uniform over the range $[10^{-4}, 10^4]$.
    \item $\gamma$: Log uniform over the range $[10^{-4}, 1]$.
\end{itemize}

\textbf{Experiment of Section \ref{sec:telvsxgb}}: 

TEL: 
\begin{itemize}
    \item Learning rate: Uniform over $\{10^{-1}$, $10^{-2}$, \dots, $10^{-5}\}$.
    \item Batch size: Uniform over $\{32, 64, 128, 256, 512\}$.
    \item Number of Epochs: Discrete uniform over $[5,100]$.
    \item $\gamma$: Log uniform over $[10^{-4}, 1]$.
    \item Tree Depth: Discrete uniform over $[2,8]$.
    \item Number of Trees: Discrete uniform over $[1,100]$.
    \item L2 Regularization for $W$: Mixture model of $0$ and the log uniform distribution over $[10^{-8}, 10^2]$. Mixture weights are $0.5$ for each.
\end{itemize}

XGBoost: 
\begin{itemize}
    \item Learning rate: Uniform over $\{10^{-1}$, $10^{-2}$, \dots, $10^{-5}\}$.
    \item Tree Depth: Discrete uniform over $[2,20]$.
    \item Number of Trees: Discrete uniform over $[1,500]$.
    \item L2 Regularization ($\lambda$): Mixture model of $0$ and the log uniform distribution over $[10^{-8}, 10^2]$. Mixture weights are $0.5$ for each.
    \item min\_child\_weight = 0.
\end{itemize}

Logistic Regression: We used Sklearn's default optimizer and increased the maximum number of iterations to $1000$. We tuned over the L2 regularization parameter ($C$): Log uniform over $[10^{-8},10^{4}]$.

CART: We used Sklearn's ``DecisionTreeClassifier'' and tuned over the depth: discrete uniform over $[2,20]$.

\textbf{Experiment of Section \ref{sec:telvdense}}: 
CNN-Dense has the following architecture:
\begin{itemize}
    \item Convolutional layer 1: has $f$ filters and a $3 \times 3$ kernel (where $f$ is a tuning parameter).
    \item Convolutional layer 2: has $2f$ filters and a $3 \times 3$ kernel.  
    \item $2 \times 2$ max pooling
    \item Flattening
    \item Dropout: the dropout rate is a tuning parameter.
    \item Batch Normalization
    \item Dense layers: a stack of ReLU-activated dense layers, where the number of layers and the units in each is a tuning parameter.
    \item Output layer: dense layer with softmax activation.
\end{itemize}

CNN-TEL has the following architecture:
\begin{itemize}
    \item Convolutional layer 1: has $f$ filters and a $3 \times 3$ kernel (where $f$ is a tuning parameter).
    \item Convolutional layer 2: has $2f$ filters and a $3 \times 3$ kernel.  
    \item $2 \times 2$ max pooling
    \item Flattening
    \item Dropout: the dropout rate is a tuning parameter. 
    \item Batch Normalization
    \item Dense layer: the number of units is a tuning parameter.
    \item TEL
    \item Output layer: softmax.
\end{itemize}

We used the following hyperparameter distributions: 
\begin{itemize}
    \item Learning rate: Uniform over $\{10^{-1}$, $10^{-2}$, \dots, $10^{-5}\}$.
    \item Batch size: Uniform over $\{32, 64, 128, 256, 512\}$.
    \item Number of Epochs: Discrete uniform over $[1,100]$.
    \item $\gamma$: Log uniform over $[10^{-4}, 1]$.
    \item Tree Depth: Discrete uniform over $[2,6]$.
    \item Number of Trees: Discrete uniform over $[1,50]$.
    \item $f$: Uniform over $\{4,8,16,32\}$.
    \item Number of dense layers in CNN-Dense: Discrete Uniform over $[1,5]$.
    \item Number of units in dense layers of CNN-Dense: Uniform over $\{16, 32, 64, 128, 256, 512\}$.
    \item Number of units in the dense layer of CNN-TEL: Uniform over $\{16, 32, 64\}$.
    \item Dropout rate: Uniform over $[0.1,0.5]$.
\end{itemize}

\end{document}